# MD-PNOP: Equation-Recast Neural Operators for Minimal-Data Extrapolation and PDE Solver Acceleration

Qiyun Cheng, Md Hossain Sahadath, Huihua Yang, Shaowu Pan, and Wei Ji*

Department of Mechanical, Aerospace, and Nuclear Engineering

Rensselaer Polytechnic Institute, 110 8th Street, Troy, NY 12180

**Abstract**

The computational overhead of traditional numerical solvers for partial differential equations (PDEs) remains a critical bottleneck for large-scale parametric studies and design optimization. We introduce a Minimal-Data Parametric Neural Operator Preconditioning (MD-PNOP) framework, which establishes a new strategy for accelerating parametric PDE solvers while strictly preserving physical constraints. To address the extrapolation limitation of neural operators, parameter-induced operator difference is recast as additional source terms and incorporated into an iterative solution scheme using a pretrained neural operator. This equation-recast formulation enables systematic parameter extrapolation from a single training configuration to a broad range of unseen parameter settings without retraining. The neural operator predictions are then embedded into iterative PDE solvers as improved initial guesses, thereby reducing convergence iterations without sacrificing accuracy. Unlike purely data-driven approaches, MD-PNOP guarantees that the governing equations remain fully enforced, eliminating concerns regarding loss of physics or interpretability. The framework is architecture-agnostic and is demonstrated using both Deep Operator Networks (DeepONet) and Fourier Neural Operators (FNO) for Boltzmann transport equation solvers in neutron transport applications. Numerical results demonstrate that neural operators trained on a single set of constant parameters successfully accelerate solutions with heterogeneous, sinusoidal, and discontinuous parameter distributions. Moreover, MD-PNOP consistently achieves approximately 50% reduction in computational time while maintaining full-order fidelity for fixed-source, single-group eigenvalue, and multigroup coupled eigenvalue problems. These results establish MD-PNOP as a robust and generalizable strategy for accelerating PDE solvers for parametric problems, balancing computational efficiency with physical accuracy.

## 1. Introduction

Partial differential equations (PDEs) provide fundamental frameworks for modeling and analyzing a wide range of scientific and engineering problems [1]. While traditional model-based numerical solvers, such as finite element, finite volume and finite difference methods, are well established and deliver high-fidelity solutions, their substantial computational cost is increasingly prohibitive with the increase in system complexity. This creates a critical bottleneck for emerging applications that demand high-throughput simulation capabilities, including large-scale parametric analysis, and iterative design optimization [2].

Recently, data-driven approaches, particularly those based on neural networks, have emerged as promising alternatives for solving PDEs [3]. Supported by the development of high-performance GPUs, well-trained neural networks can perform inference several orders of magnitude faster than conventional numerical solvers, providing a significant advantage in reducing simulation time. Among these approaches, two major categories have gained intensive attention: Physics-Informed Neural Networks (PINNs) [4] and Neural Operators [5]. PINNs directly encode the governing equations into the training process. By using automatic differentiation to calculate the derivatives, they construct a loss function that measures how well the network's output obeys the specified physics. This physics-constrained learning makes PNNNs naturally suited for parametric studies [4]. PINNs have received significant attention and have been successfully applied to various domains, such as fluid mechanics [6] and neutron diffusion [7]. Variants of PINNs have also been developed to enhance performance, such as for parametric PDEs [8] and inverse problems [9]. However, training PINNs remains challenging, and their solution accuracy generally does not yet match that achieved by traditional model-based solvers [10]. Neural operators represent another promising architecture for developing surrogate PDE solvers [5]. Unlike conventional neural networks that learn pointwise mappings between condition/solution data pairs, neural operators aim to learn a solution operator that maps entire condition or source functions to solution functions. By learning mappings between function spaces spanned by the input and output functions in the training dataset, neural operators mitigate dependence on specific training data distributions and improve generalization performance. Once trained, a neural operator can be used to predict solutions for any new condition within the training function space without retraining. Among neural operators, the Fourier Neural Operator (FNO) [11] and Deep Operator Network (DeepONet) [12] become particularly prominent. FNO and its variants have been applied to problems in multiphase flow [13], fusion plasma modeling [14], and material behavior prediction [15]. DeepONet and its variants have been employed in fields such as material fracture analysis [16], heat transfer [17], and neutron transport calculations [18]. Physics-informed loss terms can also be integrated into neural operators and form hybrid physics-informed neural operator architectures [19, 20].

* Corresponding author: jiw2@rpi.edu

Despite these advances, purely data-driven approaches continue to face critical challenges that hinder their practical deployment. These challenges include the large amounts of training data required spanning parameter space, limited robustness when extrapolating beyond the parameter regimes represented in the training set, and potential degradation in accuracy on unseen configurations. In addition, the black-box nature of such models raises concerns regarding interpretability, reliability, uncertainty quantification, and the strict enforcement of physical constraints. Hybrid approaches that integrate neural operators into conventional numerical solvers have therefore been explored as an alternative to purely data-driven PDE solvers [21-25]. In these methods, neural operators are typically trained as abstract preconditioners [21-23] or iteration-level accelerators [23-25] to improve the convergence of iterative solvers. While these approaches have demonstrated promising acceleration for fixed or moderately varying problem settings, their integration into the numerical solution process introduces additional considerations related to implementation complexity, robustness, and stability. Moreover, the generalization of such learned operators across parameter variations remains largely data-driven and is often limited by the coverage of the training data.

Motivated by the above considerations, this work introduces a minimal-data, architecture-agnostic neural operator preconditioning framework, termed MD-PNOP, that follows a fundamentally different design principle. Instead of pursuing a neural operator trained over a wide range of equation parameter space [21-25], a neural operator model trained only at a single set of parameter values is needed. The generalization over all possible parameter values is realized by recasting the governing equation at arbitrary parameter values into a form that (1) consists of extra source terms due to parameter value change from nominal values used for training to new values, and (2) allows the trained neutral operator to be iteratively employed to fast solve the recast equation. This neural operator embedded iterative procedure avoids the time-consuming numerical solution steps while generalizing to unseen configurations. To ensure high-fidelity realistic physics-constraints, the resultant solutions are then used as an initial guess to model-based numerical solvers for final solutions. As a result, the final solution remains the same accuracy as the model-based numerical solution of the original governing equations, while the overall computational cost is reduced using a solution guess that is close to the true numerical solution and provided by a fast neural operator guided solution procedure. Demonstrations on Boltzmann transport equations show that the proposed framework achieves approximately a 50% reduction in wall-clock time without compromising physical accuracy.

This paper is organized in the following way. Section 2 outlines motivation and research objectives in detail. Section 3 formulates the MD-PNOP approach, including theoretical developments for general PDEs and algorithmic designs for the integration with model-based solvers as preconditioners. Section 4 introduces two neural operator architectures employed. Section 5 demonstrates the framework on accelerating the Boltzmann transport equation solvers for neutron transport problems, covering fixed source, single-group eigenvalue, and multigroup coupled eigenvalue scenarios. Finally, Section 6 summarizes the key findings.

## 2. Motivation and Research Objectives

Despite the rapid advancements in neural network-based models for scientific and engineering applications, several critical challenges still hinder their widespread practical deployment. First, data acquisition and the associated computational cost remain major obstacles [20]. Generating high-fidelity simulation data required for training is often prohibitively time-consuming and computationally intensive. Furthermore, complex dynamic systems typically involve multiple physical or material parameters, while parameter sweeps cannot feasibly cover the entire parameter space of interest. Consequently, when using a neural network-based surrogate model, the original problem must paradoxically be solved repeatedly in advance to construct the surrogate, reducing practical efficiency. Second, maintaining solution accuracy and ensuring robust generalization present significant difficulties. Although conventional loss metrics such as mean squared error (MSE) may converge to values on the order of $10^{-7}\sim10^{-8}$, this typically corresponds to an absolute solution error of larger than the order of $10^{-4}$, which is inadequate for high-fidelity or safety-critical applications [26]. Moreover, accuracy often degrades on unseen test cases and deteriorates further when extrapolating beyond the training domain. This inability to extrapolate reliably is the most restrictive for parametric studies and design optimization of all neural network models. Lastly, the inherent "black box" nature of neural networks raises concerns regarding interpretability and physical reliability [27]. In conventional training pipelines, physical knowledge is introduced only through soft physics constraints, which are auxiliary penalty terms added to the loss function that discourage, but do not strictly prevent, violations of conservation laws and governing equations. Because of the weakly imposed constraints, the resulting surrogate offers no guarantee of physical consistency. This limitation is especially critical in design optimization and other safety-relevant decision-making tasks, where strict adherence to physical laws and model transparency are essential.

To address the above challenges, this study proposes the MD-PNOP framework, a hybrid approach combining a neural operator with the perturbation-theory-inspired equation recast and integrate it into a model-based solver. The primary objectives are to: (i) develop neural operators trained on minimal datasets, utilizing only one or a few parameter settings to substantially reduce data acquisition costs; (ii) recast governing equations inspired by perturbation theory to enable generalization across a wide parameter space beyond those used in training; and (iii) incorporate the neural operator with the recast equation into a neural operator-based preconditioner, where the prediction serves as an improved initial guess for traditional model-based solvers. This approach alleviates potential accuracy degradation and addresses interpretability concerns by rigorously enforcing physical constraints in the final solution. It is worth noting that, unlike traditional matrix preconditioning concepts, here the neural operator provides an improved initial guess for iterative PDE solvers, thereby accelerating convergence. Through this hybrid solver design, the final solution is entirely obtained from the numerical solver, strictly following governing equations and eliminating black-box concerns. Furthermore, since the final solutions are obtained from the model-based solvers, uncertainty propagation associated with input parameters can be addressed using well-established techniques developed for conventional numerical solvers.

In summary, the proposed MD-PNOP framework aims to achieve accurate, efficient, and physically consistent predictions. The framework is architecture-agnostic and is demonstrated using DeepONet and FNO to solve Boltzmann transport equations for neutron transport problems. The testing cases include fixed source problems, one-group and multigroup eigenvalue problems. The neural operators are trained only once using data generated from the transport problem with a specific set of parameter values but are then extended for solving problems with arbitrary parameter values through the MD-PNOP framework without re-training. The results show that the developed MD-PNOP framework provides a robust foundation for accelerating PDE solvers and advancing optimization in complex engineering systems.

## 3. Minimal-Data Parametric Neural Operator Preconditioning (MD-PNOP) Framework

In this section, the methodology of the MD-PNOP framework is discussed. We begin with the perturbation-theory-inspired equation recast, which reformulates parameter variations as additional source terms. This recast is the key to enabling systematic parametric generalization from minimal data, directly overcoming the fundamental extrapolation limitation of neural networks. Subsequently, we describe how a neural operator trained on minimal dataset can be generalized into new parameter settings and integrated as a preconditioner for both prescribed source and eigenvalue problems.

### 3.1 Operator Representation and Recast of PDEs for Parametric Generalization

Let $\boldsymbol{x} \in \Omega \subseteq \mathbb{R}^d$ denote the generalized coordinate vector, such as the spatial, angular, or any other relevant coordinates, $u(\boldsymbol{x})$ be the unknown solution function, and $\boldsymbol{p} \in R^m$ a vector of parameters characterizing physical constants or material properties. Let $S(\boldsymbol{x})$ denote a prescribed source term. For steady state PDEs, the equation can be written in the operator form:

$$\mathcal{L}(\boldsymbol{p})[u(\boldsymbol{x})] = S(\boldsymbol{x}) \tag{1}$$

where $\mathcal{L}(\boldsymbol{p})$ could be a linear or nonlinear operator acting on $u(\boldsymbol{x})$ and parametric dependence through $\boldsymbol{p}$. This formulation includes both the structure of the PDE, and the physical properties encoded in the parameter vector $\boldsymbol{p}$, such as cross sections, diffusion coefficients, or elastic moduli. The solution $u(\boldsymbol{x})$ is the field function of the variable of interest (e.g., temperature distribution, neutron flux), while $S(\boldsymbol{x})$ represents known external sources or forcing terms.

The exact solution of a PDE can be formally expressed using the inverse of the associated operator:

$$u(\boldsymbol{x}) = \mathcal{L}^{-1}(\boldsymbol{p})[S(\boldsymbol{x})] \tag{2}$$

where $\mathcal{L}^{-1}(\boldsymbol{p})$ denotes the corresponding solution operator that maps the source term $S(\boldsymbol{x})$ to the solution field function $u(\boldsymbol{x})$. In practice, computing this operator directly through the matrix inversion in a discretized setting is often computationally expensive and potentially ill-conditioned, particularly for high-dimensional or large-scale problems. As a result, classical numerical solvers approximate the action of the solution operator through iterative schemes. Let $u^{init}(\boldsymbol{x})$ denote an initial guess of the solution. A classical numerical solver for the PDE can be expressed in terms of a nonlinear iteration operator $\boldsymbol{\mathcal{M}}$, such that:

$$u^{(1)}(\boldsymbol{x}) = \boldsymbol{\mathcal{M}}\big(u^{(0)}(\boldsymbol{x}) = u^{init}(\boldsymbol{x}), S(\boldsymbol{x}); \boldsymbol{p}\big), \tag{3a}$$

$$u^{(l+1)}(\boldsymbol{x}) = \boldsymbol{\mathcal{M}}\big(u^{(l)}(\boldsymbol{x}), S(\boldsymbol{x}); \boldsymbol{p}\big), \qquad l = 1, 2, 3, \ldots \tag{4b}$$

where $\boldsymbol{\mathcal{M}}$ denotes the nonlinear iteration operator associated with the underlying numerical discretization and solution strategy. The numerical solver corresponds to the repeated application of $\boldsymbol{\mathcal{M}}$ until a prescribed convergence criterion is satisfied. Although such iterative procedures are robust and widely adopted, they can be computationally demanding, especially for large-scale problems or in scenarios involving parameter sweeps, inverse problems, or optimization loops where the PDE must be solved repeatedly under varying parameter configurations $\boldsymbol{p}$.

In the MD-PNOP framework, the solution operator $\mathcal{L}^{-1}(\boldsymbol{p})$ is approximated directly using a neural operator trained on a minimal dataset corresponding to only a single parameter setting $\boldsymbol{p}^*$. The trained neural operator $\mathcal{L}_{NO}^{-1}(\boldsymbol{p}^*)$ is designed to take any source term $S(\boldsymbol{x})$ and output the corresponding solution function $u_{NO}(\boldsymbol{x}; \boldsymbol{p}^*)$ for the system parameterized by $\boldsymbol{p}^*$. This output serves as an approximation of the true solution $u(\boldsymbol{x})$:

$$u_{NO}(\boldsymbol{x}; \boldsymbol{p}^*) = \mathcal{L}_{NO}^{-1}(\boldsymbol{p}^*)[S(\boldsymbol{x})] \approx \mathcal{L}^{-1}(\boldsymbol{p}^*)[S(\boldsymbol{x})] = u(\boldsymbol{x}) \tag{5}$$

Now, consider a new PDE instance with a different parameter setting $\boldsymbol{p}$ .

$$\mathcal{L}(\boldsymbol{p})[u(\boldsymbol{x})] = S(\boldsymbol{x}) \tag{5}$$

To enable the use of the trained neural operator $\mathcal{L}_{NO}^{-1}(\boldsymbol{p}^*)$ for the new system, we designed the algorithm inspired by perturbation theory. Specifically, the new parameter vector $\boldsymbol{p}$ is treated as a perturbation around the trained parameters setting $\boldsymbol{p}^*$, and decomposed as:

$$\boldsymbol{p} = \boldsymbol{p}^* + \Delta\boldsymbol{p} \tag{6a}$$

Correspondingly, the operator defined with the new parameter can be recast as:

$$\mathcal{L}(\boldsymbol{p})[u(\boldsymbol{x})] = \big(\mathcal{L}(\boldsymbol{p}^*) + \mathcal{L}(\boldsymbol{p}) - \mathcal{L}(\boldsymbol{p}^*)\big)[u(\boldsymbol{x})] = \mathcal{L}(\boldsymbol{p}^*)[u(\boldsymbol{x})] + \delta\mathcal{L}(\Delta\boldsymbol{p})[u(\boldsymbol{x})] = S(\boldsymbol{x}) \tag{6b}$$

where $\delta\mathcal{L}(\Delta\boldsymbol{p})[u(\boldsymbol{x})]$ denotes the operator difference. This leads to a recast PDE, equivalent to Eq. (5):

$$\mathcal{L}(\boldsymbol{p}^*)[u(\boldsymbol{x})] = S(\boldsymbol{x}) - \delta\mathcal{L}(\Delta\boldsymbol{p})[u(\boldsymbol{x})], \tag{7a}$$

or,

$$u(\boldsymbol{x}) = \mathcal{L}^{-1}(\boldsymbol{p}^*)\big[S(\boldsymbol{x}) - \delta\mathcal{L}(\Delta\boldsymbol{p})[u(\boldsymbol{x})]\big] \tag{6b}$$

This reformulation results in a modified PDE with an additional source term that depends on the unknown solution $u(\boldsymbol{x})$. Leveraging the pretrained neural operator $\mathcal{L}_{NO}^{-1}(\boldsymbol{p}^*)$ without retraining, the solution to Eq. (5) or Eq. (7) can be obtained by an iterative scheme with a fast neural operator predictor:

$$u^{(k+1)}(\boldsymbol{x}) = \mathcal{L}_{NO}^{-1}(\boldsymbol{p}^*)\left[S(\boldsymbol{x}) - \delta\mathcal{L}(\Delta\boldsymbol{p})\left[u^{(k)}(\boldsymbol{x})\right]\right], \ \ k = 0, 1, 2, \dots \tag{7}$$

The iteration proceeds until convergence is achieved, yielding the neural operator solution under the new parameter conditions. Table 1 gives the recast form of several representative steady state PDEs, where the parameters with stars denote the parameters used in the training data.

Table 1. Example of the recast form of representative PDEs.

| | **Diffusion Equation** |
|---|---|
| **PDE** | $-\nabla \cdot \left(D(\boldsymbol{x})\nabla u(\boldsymbol{x})\right) = S(\boldsymbol{x}), \quad \boldsymbol{p} = D(\boldsymbol{x})$ |
| **Recast** | $-\nabla \cdot \left(D^*\nabla u(\boldsymbol{x})\right) = S(\boldsymbol{x}) + \nabla(D(\boldsymbol{x}) - D^*) \cdot \nabla u(\boldsymbol{x}) + (D(\boldsymbol{x}) - D^*)\nabla^2 u(\boldsymbol{x})$ |
| | **Navier-Stokes Equation** |
| **PDE** | $(u(\boldsymbol{x}) \cdot \nabla)u(\boldsymbol{x}) - \nu(\boldsymbol{x})\nabla^2 u(\boldsymbol{x}) = -\frac{1}{\rho(x)}\nabla p(\boldsymbol{x}) + f(\boldsymbol{x}), \quad \boldsymbol{p} = \nu(\boldsymbol{x})$ |
| **Recast** | $(u(\boldsymbol{x}) \cdot \nabla)u(\boldsymbol{x}) - \nu^*\nabla^2 u(\boldsymbol{x}) = -\frac{1}{\rho(x)}\nabla p(\boldsymbol{x}) + f(\boldsymbol{x}) + (\nu(\boldsymbol{x}) - \nu^*)\nabla^2 u(\boldsymbol{x})$ |
| | **Helmholtz Equation** |
| **PDE** | $-\nabla^2 u(\boldsymbol{x}) - k^2(\boldsymbol{x})u(\boldsymbol{x}) = S(\boldsymbol{x}), \quad \boldsymbol{p} = k^2(\boldsymbol{x})$ |
| **Recast** | $-\nabla^2 u(\boldsymbol{x}) - (k^*)^2 u(\boldsymbol{x}) = S(\boldsymbol{x}) + (k^2(\boldsymbol{x}) - (k^*)^2)u(\boldsymbol{x})$ |
| | **Burger's Equation** |
| **PDE** | $u(\boldsymbol{x}) \cdot \nabla u(\boldsymbol{x}) - \nu(\boldsymbol{x})\nabla^2 u(\boldsymbol{x}) = S(\boldsymbol{x}), \quad \boldsymbol{p} = \nu(\boldsymbol{x})$ |
| **Recast** | $u(\boldsymbol{x}) \cdot \nabla u(\boldsymbol{x}) - \nu^*\nabla^2 u(\boldsymbol{x}) = S(\boldsymbol{x}) + (\nu(\boldsymbol{x}) - \nu^*)\nabla^2 u(\boldsymbol{x})$ |

As shown in Table 1, when the difference term $\delta\mathcal{L}(\Delta\boldsymbol{p})[u(\boldsymbol{x})]$ can be obtained through direct subtraction or reasonable approximations, parameter variations can be absorbed into a modified source term. This equation recast method allows a neural operator pretrained on a minimal dataset (corresponding to a single parameter setting) to generalize to new parameter configurations without requiring retraining. Although equation recast formulation itself remains exact at the continuous operator level, the evaluation of the operator difference $\delta\mathcal{L}(\Delta\boldsymbol{p})[u(\boldsymbol{x})]$ may involve numerical approximations, such as finite-difference evaluations of parameter-dependent derivatives in the N-S equation. The impact of such numerical errors is problem-dependent and primarily affects the convergence behavior of the iterative scheme, typically manifesting as slower convergence or divergence rather than silent violations of physical constraints. Compared with traditional model-based solvers, although additional computation is added to evaluate the new source terms, the most time-consuming step, the spatial sweep, is replaced by neural operator inference, which is typically very fast, thereby reducing the overall computation time for solving the PDE.

### 3.2 Preconditioning for Prescribed Source Problem

While Section 3.1 introduces the generalization of a neural operator trained on a minimal dataset to arbitrary parameter settings, two critical issues arise when applying neural network-based computational approaches to industrial and safety-critical applications. First, the accuracy of neural networks, even when trained on large datasets, typically does not match the precision achieved by high-fidelity model-based solvers, particularly when the neural network is extrapolated to new test cases beyond its training regime. Second, the black-box nature of neural networks, combined with the use of soft constraints in training loss functions and the lack of explicit uncertainty quantification for input parameters, raises concerns regarding the reliability and interpretability of neural network predictions.

To address above challenges, a hybrid solution strategy is adopted in the MD-PNOP framework. In this approach, the iterative algorithm developed in Equation 8 forms a neural operator-based preconditioner, providing an initial solution estimate for the model-based solver. This strategy accelerates convergence while ensuring that the final solution retains the full accuracy and physical consistency guaranteed by traditional numerical solvers.

First, a neural operator-based solution $u_{NO}(\boldsymbol{x})$ is obtained from the following equation by an iterative solution approach.

$$u_{NO}^{(l+1)}(\boldsymbol{x}) = \mathcal{L}_{NO}^{-1}(\boldsymbol{p}^*)\left[S(\boldsymbol{x}) - \delta\mathcal{L}(\Delta\boldsymbol{p})\left[u^{(l)}(\boldsymbol{x})\right]\right], l = 0,1,2, \dots \tag{8}$$

This solution is an approximation to the solution of the system parametrized by $\boldsymbol{p}$. Then, $u_{NO}(\boldsymbol{x})$ is used as an initial guess for the model-based solver, yielding the final solution of the PDE $u(\boldsymbol{x})$:

$$u^{(1)}(\boldsymbol{x}) = \boldsymbol{\mathcal{M}}\left(u^{(0)}(\boldsymbol{x}) = u_{NO}(\boldsymbol{x}), S(\boldsymbol{x}); \boldsymbol{p}\right) \tag{9a}$$

$$u^{(l+1)}(\boldsymbol{x}) = \boldsymbol{\mathcal{M}}\left(u^{(l)}(\boldsymbol{x}), S(\boldsymbol{x}); \boldsymbol{p}\right), l = 1,2, \dots \tag{10b}$$

where $u_{NO}(\boldsymbol{x})$ is an input to the model-based solution solver, specifically to the nonlinear iteration operator $\boldsymbol{\mathcal{M}}$, by serving as an initial guess of the final solution. Thus, the MD-PNOP framework enables a neural operator trained on a minimal dataset to accelerate the convergence of a model-based solver for arbitrary new parameter settings, while ensuring that the final solution remains fully accurate and physically consistent. The key concepts of the MD-PNOP framework are summarized in Figure 1. In this framework, the prediction of the neural operator-based preconditioner is used only as an initial guess, and the final solution is entirely generated by the traditional model-based solver. The neural operator does not modify the governing equations, the spatial discretization, or the iteration operator of the classical solver. Therefore, all established parameter uncertainty propagation and sensitivity analysis techniques developed for conventional model-based solvers remain directly applicable. Since the final solution strictly satisfies the original governing equations, MD-PNOP delivers a fully physics-constrained and high-accuracy solution while reducing the overall computational cost.

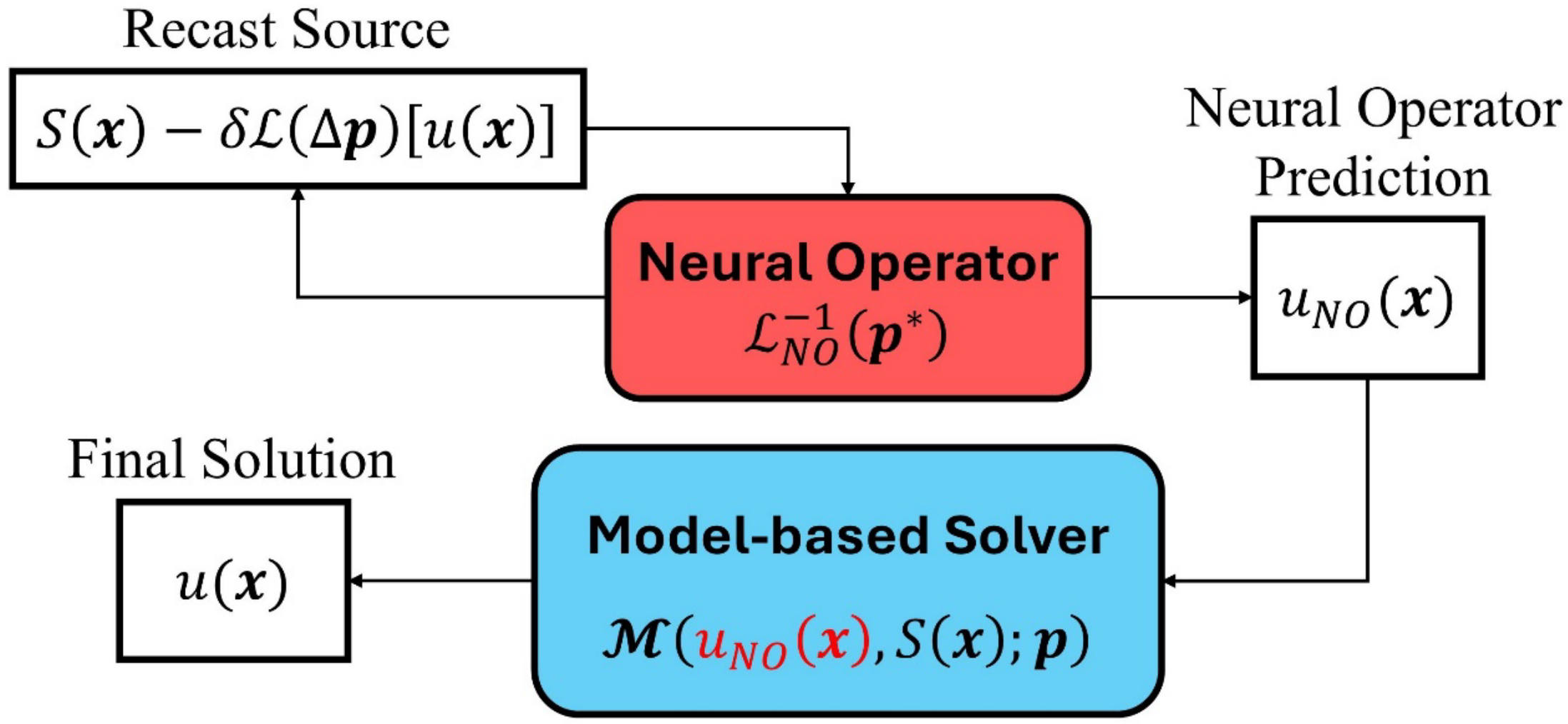

Figure 1. Key concepts of the MD-PNOP framework. The equation recast treats parameter deviations between the trained neural operator and the current problem as additional source terms, enabling accurate generalization without retraining. Neural operator predictions are then used as improved initial guesses for the iterative model-based solver, which refines the solution to guarantee full-order accuracy while remaining strictly constrained by the governing physics.

Hybrid approaches that integrate neural networks or neural operators into conventional numerical solvers have been actively investigated as alternatives to purely data-driven PDE solvers. Existing neural preconditioning strategies typically follow two directions: (i) approximating preconditioners for the discretized system, such as inverse mappings or residual corrections [21-23], and (ii) embedding learned operators within the solver loop to generate iterative updates [23-25]. While these approaches can provide acceleration, their robustness under parameter variations often depends on the coverage of the training data and may introduce additional stability considerations due to modifications within the iteration process. In contrast, MD-PNOP achieves parameter generalization through an equation-recast formulation that explicitly transforms parameter deviations into additional source terms. The neural operator is invoked only to generate an improved initial iterate, and all subsequent iterations are performed by the original model-based solver. Therefore, the iteration operator and stability properties of the underlying numerical scheme remain unchanged, while computational cost is reduced through faster convergence.

### 3.3 Discussion on the scope and applicability of MD-PNOP

High-fidelity numerical solvers remain the most reliable tools for solving PDEs, but their computational cost can be prohibitive for large-scale parameter scans, optimization, or real-time applications. At the same time, purely data-driven neural network models, including neural operators, generally exhibit limited reliability when extrapolated beyond their training distributions. In safety-critical applications such as reactor neutronics, physical consistency and solution fidelity take precedence over raw inference speed. The MD-PNOP framework is therefore designed to prioritize strict enforcement of governing equations and solution accuracy, while reducing computational cost through a preconditioning strategy.

Several approaches have been proposed to improve parameter generalization in neural network based PDE solvers. One class of methods incorporates physics-informed loss terms [28], as in PINNs [5], PI-DeepONets [29], and PINOs [30], where governing equations are embedded into the training objective via automatic differentiation. These methods can significantly reduce data requirements, in some cases even eliminating the need for labeled data and are well suited for learning parametric PDEs. Another class of approaches relies on transfer learning [31] or meta-learning [32] strategies, in which neural networks are pretrained across related tasks and efficiently adapted to new parameter configurations using limited additional data. Such methods have shown strong performance when different tasks share similar data distributions or latent structures.

MD-PNOP follows a different design philosophy for parameter extrapolation. Instead of relying on loss-based soft regularization or data-driven task transfer to implicitly learn extrapolative behavior, it adopts an equation recast formulation in which parameter variations are explicitly represented as additional source terms in the governing equations. This formulation preserves the operator structure of the PDE and enables systematic parameter extrapolation independent of the underlying neural network architecture. While the resulting iterative scheme may limit the achievable acceleration compared to purely data-driven inference, it prioritizes solution fidelity, robustness, and reliable extrapolation to unseen parameter configurations under hard physical constraints. In this sense, the equation-recast strategy employed in MD-PNOP complements existing parameter generalization approaches by emphasizing physics consistency rather than data-driven extrapolation. A conceptual comparison among representative approaches is summarized in Table 2.

Table 2. Conceptual comparison of representative learning-based approaches for parameter generalization in PDE solvers

| Category | Representative Methods | Parameter Generalization Approach | Use of Governing Equation | Extrapolation Guarantee |
| --- | --- | --- | --- | --- |
| Physics-informed learning | PINNs [5]<br>PI-DeepONet [29]<br>PINO [30] | Loss-based regularization using PDE residuals | Yes, as soft constraints embedded in training loss | Data-driven, limited beyond training regime |
| Transfer/Meta learning | Transfer NO [31]<br>Meta NO [32] | Task-level knowledge transfer with limited fine-tuning | Implicit or indirect | Data-driven, task-dependent |
| **Equation recast** | **MD-PNOP (This work)** | Equation recast with explicit source-term correction | Yes, exact enforcement through operator difference | Physics-consistent extrapolation under operator continuity |

The applicability of the MD-PNOP framework is subject to several assumptions. First, the governing PDEs must be known, as the equation-recast formulation requires explicit access to the operator structure. Accordingly, the proposed framework is not intended for equation discovery or fully data-driven modeling in the absence of prior knowledge of the governing equations. Second, the approach assumes a certain degree of continuity of the solution operator with respect to the parameter space. In practical terms, this means that parameter deviations should not induce qualitative changes in the solution structure, such as bifurcations, resonance phenomena, or other singular behaviors. Since the equation-recast formulation represents parameter deviations as additional source terms, its effectiveness depends on the stability of the underlying operator under such deviations. Consequently, naive extrapolation across parameter regimes associated with singularities or abrupt structural transitions may not be appropriate. Within these bounds, MD-PNOP is well suited for accelerating existing high-fidelity solvers in parametric studies, particularly for parameter variations around baseline configurations, and for serving as a physics-constrained surrogate in digital twin applications.

As discussed in Section 3.1, the practical evaluation of the operator difference, especially for high-dimensional operators or terms involving high-order derivatives, may rely on numerical approximations and can become challenging for complex systems. However, the equation-recast framework does not require all parameters to be recast. Selective or partial recasting of parameter dependencies can still be employed to reduce the dimensionality of the training dataset while simultaneously lowering the complexity of evaluating the operator difference. This flexibility provides a principled trade-off among approximation accuracy, training cost, and robustness, and offers an effective mechanism for controlling the impact of operator-difference approximation errors in practical applications.

Finally, it is emphasized that the MD-PNOP framework is agnostic to the choice of neural operator architecture. In this work, both DeepONet and FNO backbones are employed solely to demonstrate this architecture-agnostic property. Neural operators are trained to achieve comparable levels of training loss without architecture-specific fine-tuning or extensive hyperparameter optimization, as the objective is not to benchmark neural operator performance but to evaluate the robustness and effectiveness of the MD-PNOP framework itself. As advances in neural operator architecture continue, the performance of MD-PNOP is expected to further improve accordingly.

## 4. Neural Operator Architectures

An operator is a mathematical entity which, when applied to a function, produces another function [1]. Recently, neural operators have gained significant attention for solving PDEs and have demonstrated superior generalization performance compared to traditional neural networks [5]. Serving as surrogates for analytical solution operators, neural operators represent a new class of neural networks that directly map input functions, such as initial conditions, boundary conditions, or source distributions, to solution functions. By learning a mapping between infinite-dimensional function spaces, a well-trained neural operator can handle any new input condition or source distribution function within the function space spanned by the training dataset, without requiring retraining. In this section, two widely used neural operator architectures are briefly introduced: DeepONet and FNO. Both architectures are integrated into our MD-PNOP framework to demonstrate the proposed algorithms and to highlight the framework's independence from specific neural operator architectures.

### 4.1 Deep Operator Network (DeepONet)

The DeepONet can be viewed as an analogy to the proper orthogonal decomposition (POD) approach [33]. As illustrated in Figure 2(a), the DeepONet architecture consists of two subnetworks: a trunk network (TN) and a branch network (BN). TN learns functional basis modes from the training dataset, with the number of modes truncated by the number of neurons in its output layer. Once trained, the TN takes a position vector $\boldsymbol{y}$ from the output domain, such as a time point or a spatial coordinates pair, and outputs mode values $\boldsymbol{N_T}$ representing the physical behavior at the given point $\boldsymbol{y}$. The BN, on the other hand, takes discretized condition functions $[u(x_1) \quad ... \quad u(x)_n]$ such as boundary or initial conditions, as inputs, and provides coefficients $\boldsymbol{N_B}$ corresponding to the different basis modes under the given condition. The final output of the DeepONet is the inner product of the TN and BN outputs, approximating the target function $G$ at given position $\boldsymbol{y}$ under condition $u$:

$$G(u)(y) \approx \langle N_T, N_B \rangle \tag{11}$$

This design allows the DeepONet to rapidly predict solutions for new inputs across different conditions and scenarios. Moreover, since the TN can accept any continuous coordinate vector in the output domain, DeepONet is mesh-independent and effectively represents continuous functions.

DeepONet has been successfully applied in various domains, including material fracture prediction [16], heat transfer problems [17], neutron transport calculation [18], and neutron shielding analysis [19]. Variants such as the Proper Orthogonal Decomposition-DeepONet (POD-DeepONet) [33] and the Singular Value Decomposition-DeepONet (SVD-DeepONet) [34] have been developed to further enhance performance across diverse applications.

**4.2 Fourier Neural Operator (FNO)**

The FNO is inspired by spectral methods for solving PDEs [11], and its architecture is shown in Figure 2(b). The FNO first takes the discretized condition function $[u(x_1) \quad \dots \quad u(x)_n]$ and lifts it into a higher-dimensional representation $\tilde{u}$ using a pointwise shallow fully connected neural network. The lifted function $\tilde{u}$ is then processed through multiple Fourier layers. In each Fourier layer, $\tilde{u}$ follows two parallel paths:

- In the first path, a fast Fourier transform (FFT) is applied to $\tilde{u}$, followed by a linear transformation in Fourier space to filter or truncate high-frequency Fourier modes. The filtered Fourier modes are then mapped back to the spatial domain using an inverse FFT (IFFT).
- In the second path, $\tilde{u}$ is processed through a pointwise linear transformation in the spatial domain.

The outputs from these two paths are summed and then passed through a nonlinear activation function $\sigma$, introducing nonlinearity into the operator. After passing through $N$ Fourier layers, the high-dimensional representation $\tilde{u}_N$ is projected back to the target dimension to produce the final prediction of the FNO. By learning the mapping in the global Fourier space, the FNO effectively captures long-range dependencies and continuous functional relationships, making it particularly suitable for problems involving continuous fields. Additionally, since the mapping is learned in the spectral domain, the trained FNO is inherently mesh-invariant and can be applied to finer spatial resolutions at inference, enabling zero-shot super-resolution capabilities [11].

The Fourier Neural Operator (FNO) has been successfully applied to plasma modeling [14], elastic wave simulations [35], and urban microclimate predictions [36]. Furthermore, variants, such as implicit FNO (IFNO) [15], and factorized FNO (F-FNO) [35] have been developed to enhance performance in specific domain applications. The FNO is also embedded into hybrid frameworks to improve the performance of low order numerical solvers [37].

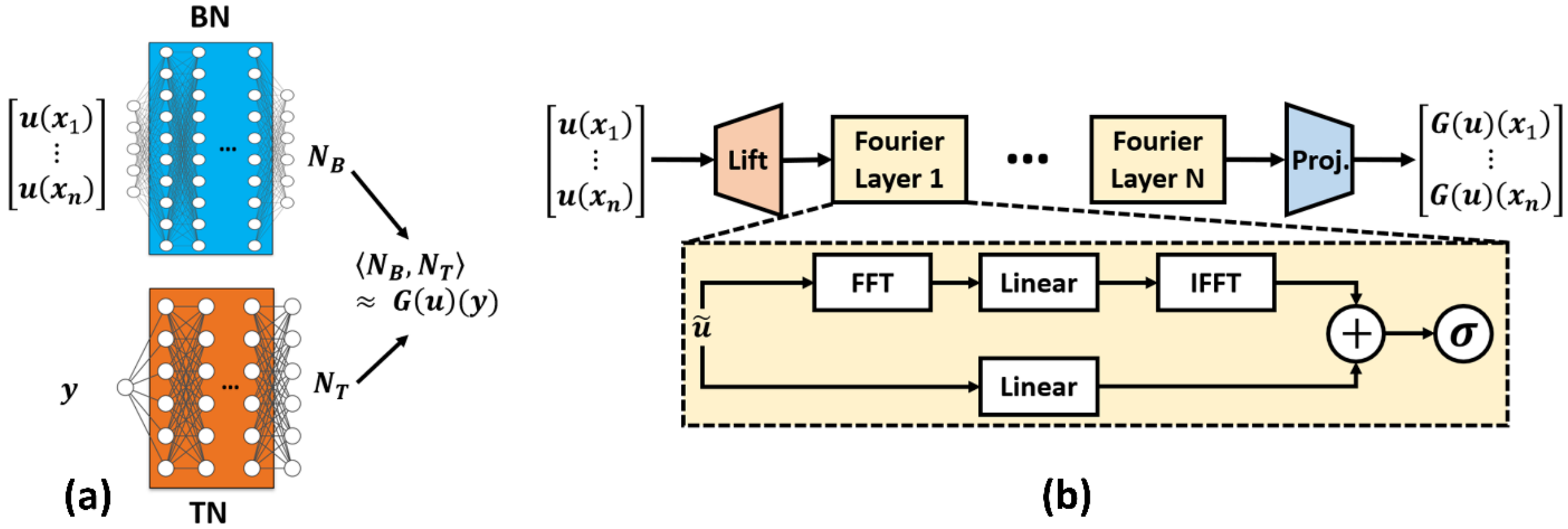

Figure 2: Architectures of (a) the Deep Operator Network and (b) the Fourier Neural Operator.

## 5. Case Study: Neutron Transport Equation

In this section, the proposed MD-PNOP framework is demonstrated using the one-dimensional neutron transport equation (NTE). The NTE, as a specific form of the Boltzmann transport equation, plays a crucial role in various scientific and engineering applications, including nuclear reactor design, radiation shielding, and material irradiation analysis. Moreover, the fixed source problem, the single-group eigenvalue problem, and the multigroup eigenvalue problem based on the NTE are well-studied and serve as excellent benchmarks to demonstrate the capabilities and generalization performance of the MD-PNOP framework.

### 5.1 Fixed Source Problem

The fixed source problem serves as a fundamental test case for the NTE: given a neutron source with a space-dependent neutron generation rate density, compute the steady-state neutron distribution in the domain. Typical applications of fixed source problems include neutron shielding [38], and chain reaction startups in nuclear reactors [39].

Consider a one-dimensional heterogeneous slab geometry with length $L = 10cm$, the steady-state, one-dimensional NTE with isotropic scattering and an isotropic source is written as:

$$\mu \frac{\partial \psi(x,\mu)}{\partial x} + \Sigma_t(x)\psi(x,\mu) = \frac{1}{2}\Sigma_{s,0}(x) \int_{-1}^{1} \psi(x,\mu) d\mu + \frac{1}{2} S(x) \tag{12}$$

where $\psi(x,\mu)$ is the angular neutron flux, and $\mu \in [-1,1]$ is the cosine of the angle. $\Sigma_t(x)$ is the total cross section, and $\Sigma_{s,0}(x)$ is the zeroth Legendre moment of the differential scattering cross section, and $S(x)$ is an arbitrary isotropic neutron source. For better scalability, the source term $S(x)$ is often normalized to represent a single neutron distributed over the spatial domain:

$$\int_0^L S(x) dx = 1 \tag{13}$$

Rearranging Equation 12 gives:

$$2\left(\mu \frac{\partial}{\partial x} + \Sigma_t(x) - \frac{1}{2}\Sigma_{s,0}(x) \int_{-1}^{1} d\mu \right) \psi(x,\mu) = S(x) \tag{14}$$

Thus, for the fixed source problem, the operator for the angular neutron flux can be defined as:

$$\mathcal{L}_{\psi}\left(\Sigma_t(x), \Sigma_{s,0}(x)\right) = 2\left(\mu\frac{\partial}{\partial x} + \Sigma_t(x) - \frac{1}{2}\Sigma_{s,0}(x)\int_{-1}^{1} d\mu\right) \tag{15}$$

where the operator $\mathcal{L}_{\psi}(\boldsymbol{p})$ is defined on both $x$ and $\mu$ domains, with parameter vector $\boldsymbol{p}$ comprising $\Sigma_t(x)$ and $\Sigma_{s,0}(x)$. Consequently, the angular flux solution can be expressed as:

$$\psi(x,\mu) = \mathcal{L}_{\psi}^{-1}\left(\Sigma_t(x), \Sigma_{s,0}(x)\right)[S(x)] \tag{16}$$

The scalar neutron flux $\phi(x)$ is more commonly used in engineering applications, which is obtained by integrating the angular flux over all angles:

$$\begin{aligned}\phi(x) = \int_{-1}^{1}\psi(x,\mu)d\mu = \int_{-1}^{1}\mathcal{L}_{\psi}^{-1}\left(\Sigma_t(x), \Sigma_{s,0}(x)\right)[S(x)]d\mu \\ = \left(\int_{-1}^{1}\mathcal{L}_{\psi}^{-1}\left(\Sigma_t(x), \Sigma_{s,0}(x)\right)d\mu\right)[S(x)]\end{aligned} \tag{17}$$

Therefore, we can define a solution operator for the scalar flux:

$$\mathcal{L}_{\phi}^{-1}(\Sigma_t(x), \Sigma_{s,0}(x)) = \int_{-1}^{1}\mathcal{L}_{\psi}^{-1}\left(\Sigma_t(x), \Sigma_{s,0}(x)\right)d\mu \tag{18}$$

The variable $\mu$ disappears after angular integration, and the resulting scaler flux solution operator $\mathcal{L}_{\phi}^{-1}$ can be applied to any isotropic neutron source $S(x)$. The final solution is the scalar flux $\phi(x)$ corresponding to the system of parameters $\Sigma_t(x)$, and $\Sigma_{s,0}(x)$.

In practice, model-based numerical solvers for fixed source neutron transport problems usually use the Discrete Ordinates ($S_N$) method [40]. The transport sweeps performed by the $S_N$ method across the spatial domain account for a significant portion of the total computational cost. During the design and optimization of nuclear systems, repeated adjustment of the cross sections (e.g., through modification of material compositions) is a key process to achieve desired performance objectives. Consequently, the fixed source NTE must often be solved repeatedly within the same geometry but for varying parameter sets. To accelerate model-based solvers for fixed source problems, the MD-PNOP framework integrates a neural operator trained to approximate the scalar flux solution operator defined in Equation 18 and then combines it with the recast equation as a preconditioner for the model-based solver.

To mitigate the computational cost associated with generating large training datasets, to account for the limited availability of experimental data in practical scenarios, and to emphasize the generalization capability of the proposed algorithm, the neural operator is trained using a minimal dataset corresponding to a single pair of constant cross section values, denoted as $\Sigma_t^* = 1.0cm^{-1}$ and $\Sigma_{s,0}^* = 0.5cm^{-1}$. 1000 different neutron source distributions are sampled using a Gaussian random field and each is normalized to represent a single neutron source. The corresponding scalar flux solutions are obtained via a model-based solver. Thus, the training dataset consists of 1000 samples, each with a different source distribution but with the same total and scattering cross sections. A DeepONet and a FNO are each trained on this dataset, achieving final training losses on the order of $10^{-8}$, which corresponds to an expected prediction accuracy of approximately $10^{-4}$. The same pretrained DeepONet and FNO are used across all test cases, including the eigenvalue problems presented in the following sections, without any retraining or additional fine-tuning. Additional details regarding the architectures and hyperparameters of the two neural operators are provided in Appendix A1. The trained neural operator is expected to yield scalar flux solutions for arbitrary isotropic source terms:

$$\phi_{NO}(x) = \mathcal{L}_{NO}^{-1}\left(x; \Sigma_t^*, \Sigma_{s,0}^*\right)\left(S(x)\right) \approx \mathcal{L}_{\phi}^{-1}\left(x; \Sigma_t^*, \Sigma_{s,0}^*\right)[S(x)] \tag{19}$$

To ensure final solution's accuracy, the MD-PNOP framework uses the neural operator output as the initial guess for a model-based solver, which subsequently refines the solution. The final scalar flux is given by:

$$\phi(x) = \mathcal{M}\left(\phi_{NO}(x); S(x), \Sigma_t^*, \Sigma_{s,0}^*\right) \tag{20}$$

Where $\mathcal{M}(\cdot)$ denotes the model-based solver (source iteration based on the $S_N$ method), initialized by the neural operator solution $\phi_{NO}(x)$.

To extend the applicability of the trained neural operator to more general scenarios, the neutron transport equation is recast by decomposing the spatially varying cross sections into the trained values and their deviations:

$$\begin{aligned} &\mu \frac{\partial \psi(x,\mu)}{\partial x} + (\Sigma_t(x) - \Sigma_t^* + \Sigma_t^*)\psi(x,\mu) \\ &\qquad = \frac{1}{2}\left(\Sigma_{s,0}(x) - \Sigma_{s,0}^* + \Sigma_{s,0}^*\right)\int_{-1}^{1} \psi(x,\mu)d\mu + \frac{1}{2}S(x) \end{aligned} \tag{21}$$

where $\Sigma_t(x)$ and $\Sigma_{s,0}(x)$ represent the general spatial dependent total cross section and the zeroth Legendre moment of the differential scattering cross section, respectively. Rearranging Equation 21 yields:

$$\begin{aligned} &\mu \frac{\partial \psi(x,\mu)}{\partial x} + \Sigma_t^* \psi(x,\mu) - \frac{1}{2}\Sigma_{s,0}^* \int_{-1}^{1} \psi(x,\mu)d\mu \\ &\qquad = \frac{1}{2}\left(\Sigma_{s,0}(x) - \Sigma_{s,0}^*\right)\phi(x) - (\Sigma_t(x) - \Sigma_t^*)\psi(x,\mu) + \frac{1}{2}S(x) \end{aligned} \tag{22}$$

In this formulation, the deviations between the spatially varying cross sections and the trained reference values are treated as additional source terms. To enable direct application of the pretrained neural operator, the angular flux in the additional absorption source is approximated isotopically as:

$$(\Sigma_t(x) - \Sigma_t^*)\psi(x,\mu) \approx \frac{1}{2}(\Sigma_t(x) - \Sigma_t^*)\phi(x) \tag{23}$$

This approximation arises from the choice to predict the scalar flux $\phi(x)$ directly rather than the angular flux $\psi(x,\mu)$. The MD-PNOP framework itself does not impose any fundamental restriction on this choice, however, the scalar flux formulation is adopted here due to its greater relevance in industrial applications and to demonstrate the effectiveness of the preconditioning design. The error introduced by the isotropic approximation is primarily associated with anisotropic scattering effects, but it is subsequently corrected by the model-based solver, ensuring that the final solution retains full-order accuracy and strict physical consistency.

Following the algorithm developed in Section 3, the scalar flux solution using the recast equation is updated as:

$$\phi_{NO}^{(k+1)}(x) = \mathcal{L}_{NO}^{-1}(\boldsymbol{p}^*)\left[S(x) + \left(\left(\Sigma_{s,0}(x) - \Sigma_{s,0}^*\right) - \left(\Sigma_t(x) - \Sigma_t^*\right)\right)\phi_{NO}^{(k)}(x)\right] \tag{24}$$

In this form, the right-hand side includes a modified source term compatible with the neural operator trained at $\left(\Sigma_t^*, \Sigma_{s,0}^*\right)$, and the scalar flux is solved iteratively until convergence. After convergence of the iterative scheme, the solution from the neural operator is subsequently passed to the model-based solver to obtain the final, fully accurate solution without approximation errors.

The MD-PNOP framework is evaluated using three different test cases, with both DeepONet and FNO architectures. A model-based solver employing the $S_N$ method with $N = 32$ angular discretization and spatial resolution of $dL = 0.1cm$ is used as the benchmark solver. To comprehensively demonstrate the effectiveness of the hybrid algorithm design in MD-PNOP, the following approaches are compared across all three test cases: the model-based solver alone, standalone DeepONet and FNO solvers with equation reformulation, and the corresponding hybrid solvers (denoted as DON-Pre and FNO-Pre), which further refine the neural operator outputs using the model-based solver. For fair comparison, the convergence criterion is set to $10^{-4}$ consistently across all solvers to assess simulation time and solution accuracy. The performance of the different solvers is summarized in Table 3 and Table 4.

Case 1 uses the same cross section parameters as the training dataset ($\Sigma_t^* = 1.0cm^{-1}$, $\Sigma_{s,0}^* = 0.5cm^{-1}$) but with a new source sampled from a different GRF than the one used for training. As shown in Figure 3, although both DeepONet and FNO successfully capture the general shape of the scalar flux, discrepancies remain across the spatial domain. These inaccuracies are also evident in Table 4, where the DeepONet and FNO exhibit errors on the order of $10^{-3}$ compared to the benchmark solution. This observation supports the earlier discussion on accuracy degradation in neural network-based solvers, especially when applied directly without correction. However, the inference time of the trained neural operators is extremely fast. In particular, the DeepONet benefits from its two fully connected subnetworks, resulting in inference times less than 1% of the model-based solver's runtime. By contrast, the FNO requires approximately 15.5% of the model-based solver time due to the computational overhead of FFT and IFFT operations. When integrated into the MD-PNOP hybrid framework, as shown in both Figure 3 and Table 4, the accuracy of DON-Pre and FNO-Pre reaches the expected level of $10^{-4}$, while still reducing the overall simulation time by up to 60%.

Case 2 not only modifies the cross section parameters to $\Sigma_t = 1.2cm^{-1}$, and $\Sigma_{s,0} = 0.8cm^{-1}$, with a newly sampled GRF source, but also introduces an anisotropic scattering term, which is not included in the training dataset:

$$\mu \frac{\partial \psi(x,\mu)}{\partial x} + \Sigma_t(x)\psi(x,\mu) = \frac{1}{2}\Sigma_{s,0}(x)\phi(x) + \frac{3}{2}\Sigma_{s,1}(x)\mu \int_{-1}^{1} \mu\psi(x,\mu)d\mu + \frac{1}{2}S(x) \tag{25}$$

Here, the first Legendre moment of the differential scattering cross section is set to $\Sigma_{s,1} = 0.3cm^{-1}$. Since the anisotropic scattering term is angular dependent, the neural operators trained only on isotropic scalar flux data cannot directly resolve this term using the equation reformulation strategy. However, the hybrid solver design in the MD-PNOP framework effectively addresses this limitation. As shown in Figure 4 and Table 4, due to the absence of anisotropic scattering in training, standalone DeepONet and FNO solvers exhibit noticeable inaccuracies in parts of the spatial domain, with $L_2$-norm errors degrading into the order of $10^{-2}$. After refinement by the model-based solver, however, the DON-Pre and FNO-Pre solutions achieve the expected accuracy level on the order of $10^{-4}$, while still reducing overall computational time by up to 50%.

Case 3 considers the most complex scenario, involving heterogeneous cross sections with an anisotropic scattering term. The total cross section $\Sigma_t(x)$ and the zeroth moment of the scattering cross section $\Sigma_{s,0}(x)$ are defined as sinusoidal functions, as shown in Figure 5(b), to mimic lattice-like structural variations commonly found in nuclear reactor designs. The first moment of the scattering cross section is set to $\Sigma_{s,1} = 0.3cm^{-1}$. As shown in Figure 5(a) and Table 4, solvers using the MD-PNOP framework achieve an accuracy on the order of $10^{-4}$ when implemented with both DeepONet and FNO architectures. Meanwhile, the overall computational cost is reduced by up to 35%.

Based on the above testing cases, the MD-PNOP framework demonstrates robust performance for the fixed source problem across a range of scenarios. Although the neural operators are trained on a minimal dataset, the equation reformulation technique enables the trained models to generalize effectively to all parameter configurations, including heterogeneous cross sections. The hybrid design eliminates approximation errors and converts the originally black-box neural networks into fully physics-constrained model-based solvers. Meanwhile, the overall computational cost is significantly reduced. The implementation of both DeepONet and FNO further highlights the architecture-agnostic capabilities of the MD-PNOP framework, allowing it to benefit from advancements in neural operator development. It is worth noting that the fixed source problem in a small spatial domain is already relatively fast to solve using traditional model-based solvers. As a result, the additional FFT and IFFT operations in FNO make the inference time comparable to that of the model-based solver in this specific context.

Table 3. Normalized simulation time for the fixed source problems (%).

| | Model-based | DeepONet | DON-Pre | FNO | FNO-Pre |
|---|---|---|---|---|---|
| Case 1 | 100 | 0.634 | 39.516 | 15.525 | 56.131 |
| Case 2 | 100 | 1.746 | 50.160 | 33.544 | 70.993 |
| Case 3 | 100 | 2.631 | 64.476 | 38.599 | 81.290 |

Table 4. $L_2$-norm error compared to the benchmark ($\times 10^{-4}$).

| | Model-based | DeepONet | DON-Pre | FNO | FNO-Pre |
|---|---|---|---|---|---|
| Case 1 | / | 90.1 | 4.4 | 83.9 | 1.5 |
| Case 2 | / | 161.9 | 8.0 | 108.6 | 4.6 |
| Case 3 | / | 274.9 | 6.3 | 147.8 | 6.9 |

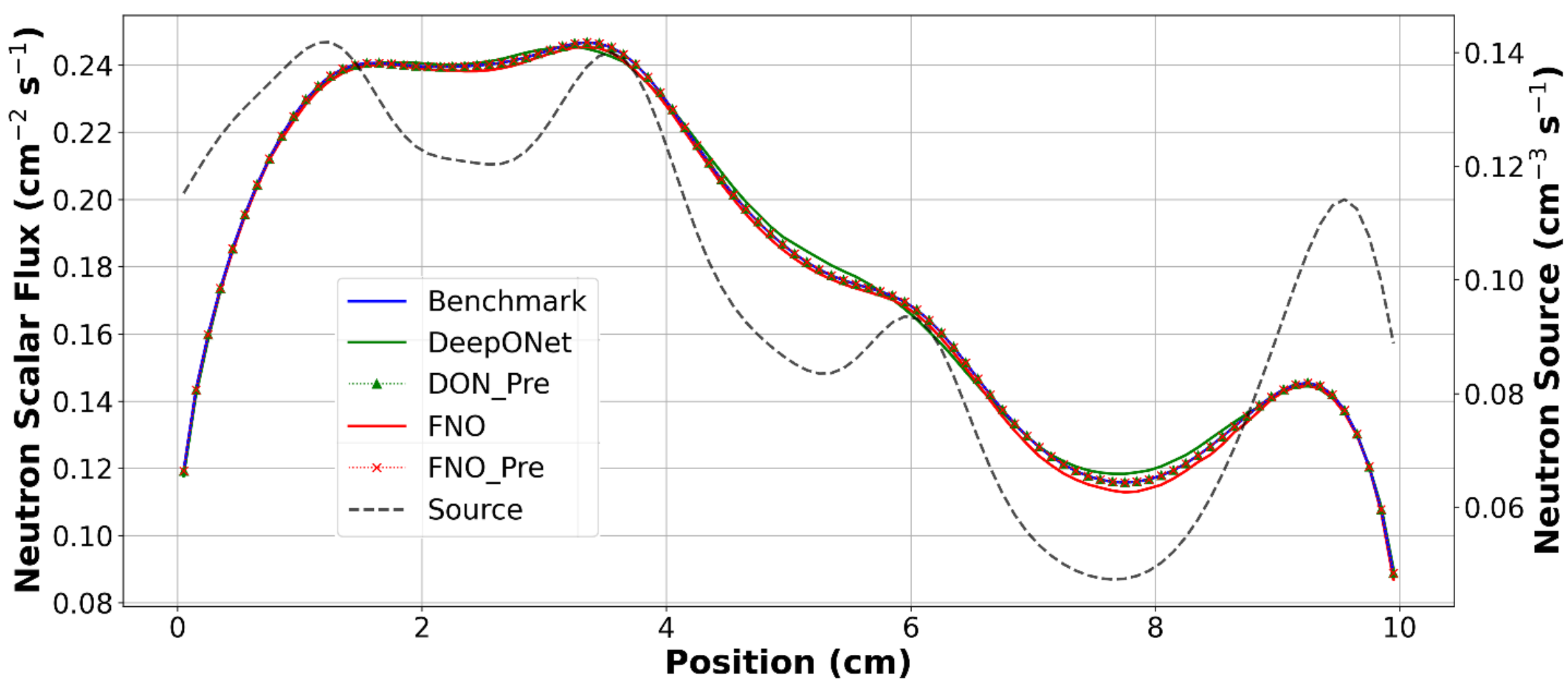

Figure 3: Neutron scalar flux comparison of the testing Case 1 for the fixed source problem. The parameters are set as $\Sigma_t^* = 1.0cm^{-1}$, and $\Sigma_{s,0}^* = 0.5cm^{-1}$, and the source is sampled from GRF.

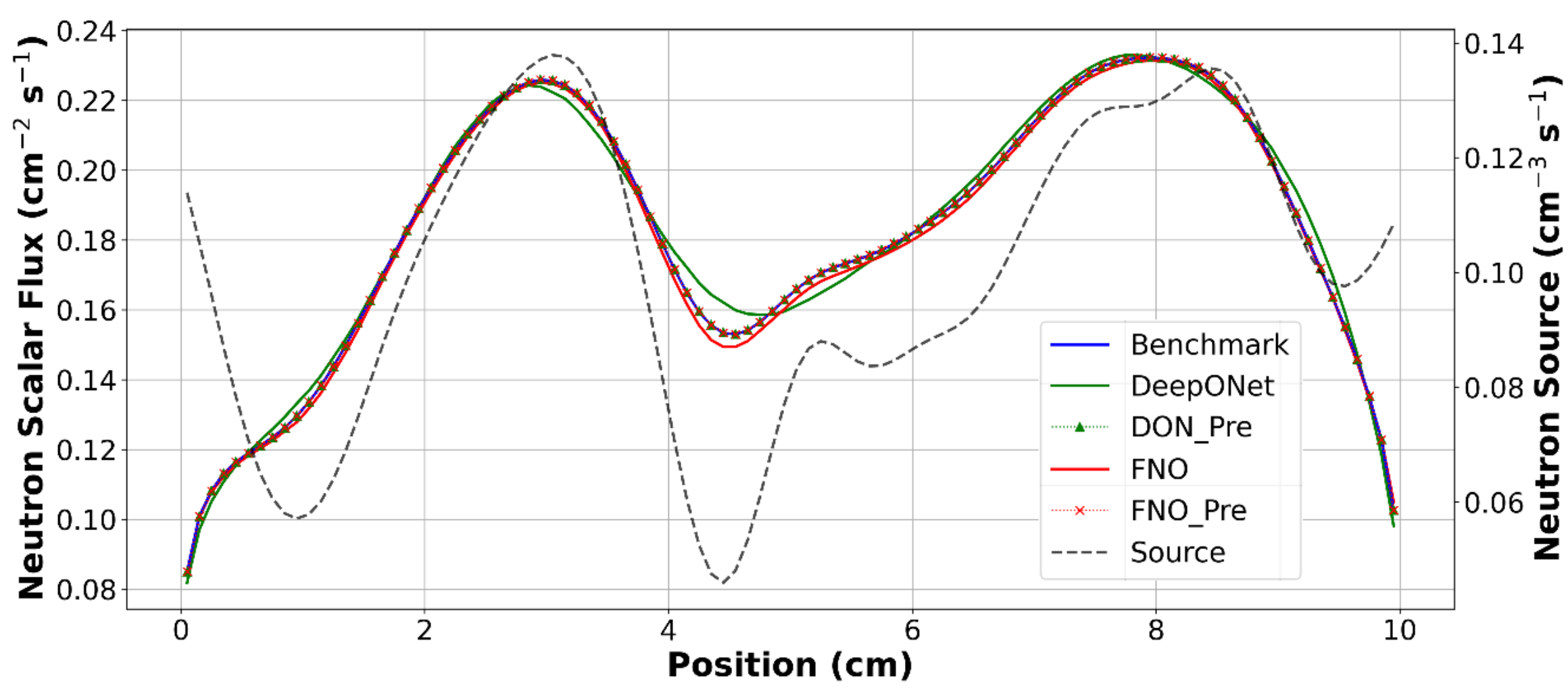

Figure 4: Neutron scalar flux comparison of the testing Case 2 for the fixed source problem including the anisotropic scattering. The parameters are set as $\Sigma_t^* = 1.2cm^{-1}$, $\Sigma_{s,0}^* = 0.8cm^{-1}$, and $\Sigma_{s,1} = 0.3cm^{-1}$. The source is sampled from GRF.

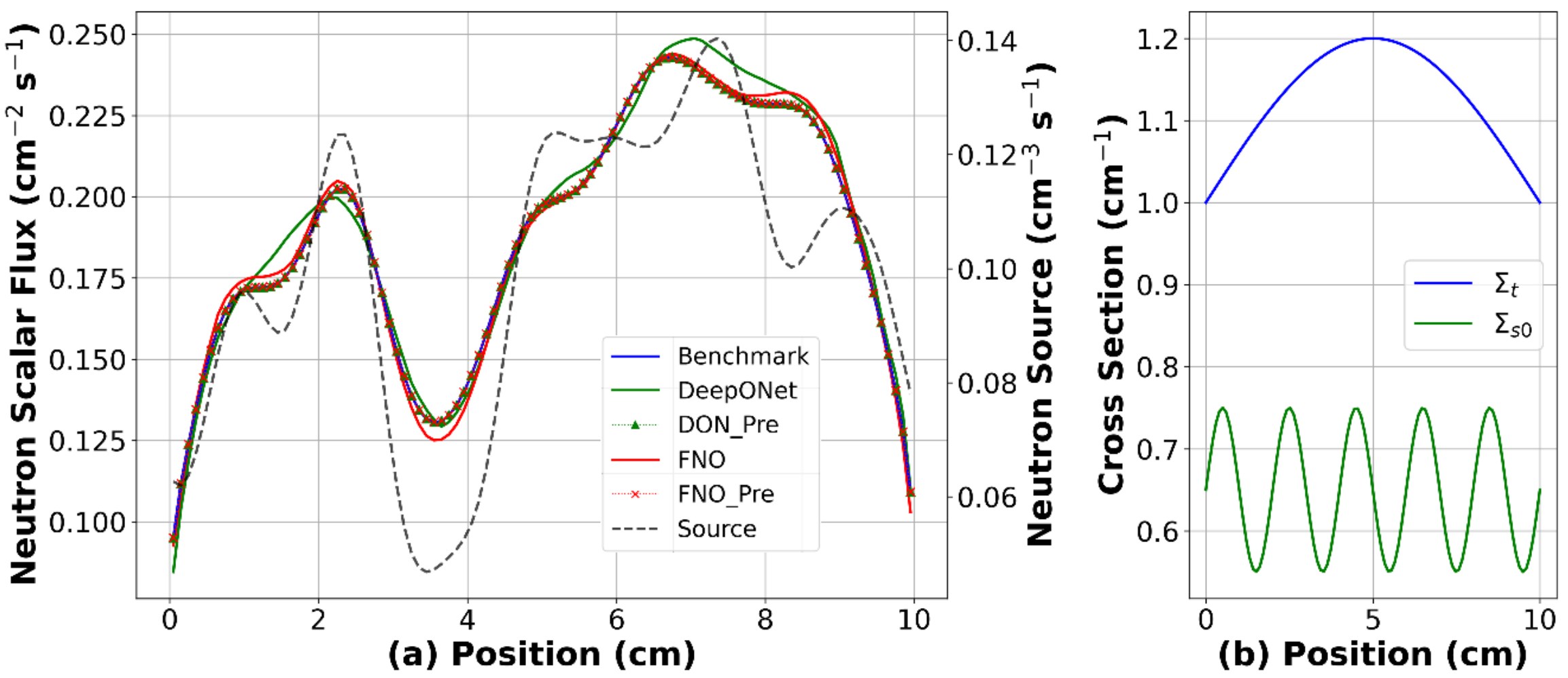

Figure 5: Testing Case 3 for the fixed source problem with heterogenous cross sections and anisotropic scattering $\left(\Sigma_{s,1} = 0.3cm^{-1}\right)$. (a) Neutron scalar flux comparison of different solvers. (b) The heterogenous cross sections $\Sigma_t(x)$ and $\Sigma_{s,0}(x)$ .

### 5.2 Eigenvalue Problem

In the context of solving neutron transport equations, calculating the eigenvalue $k$, which is also referred to as the effective neutron multiplication factor, is a fundamental step in criticality calculations and nuclear reactor design [41]. In the eigenvalue problem, the fixed external source in the standard transport equation is replaced by a flux-dependent fission source:

$$\mu \frac{\partial \psi(x,\mu)}{\partial x} + \Sigma_t(x)\psi(x,\mu) = \frac{1}{2}\Sigma_{s,0}(x)\int_{-1}^{1}\psi(x,\mu)d\mu + \frac{\nu\Sigma_f(x)}{2k}\phi(x) \tag{26}$$

where $\Sigma_f(x)$ is the spatially dependent fission cross section, and $\nu$ is the average number of neutrons produced per fission event.

When using a model-based solver, solving an eigenvalue problem typically involves two nested iterative schemes: an inner iterative solver $\boldsymbol{\mathcal{M}}$, which solves the prescribed source problem using the current eigenvalue and eigenfunction estimates, and an outer iteration $\boldsymbol{\mathcal{P}}$, which calculates the eigenvalue, normalizes the eigenfunction, and then updates the source term. The fission source is initialized from an initial guess of the scalar flux $\phi^{(init)}(x)$, and the problem is then treated as a fixed source problem. The $S_N$ method is again used to solve this fixed source problem as the inner loop $\boldsymbol{\mathcal{M}}$. In the outer loop $\boldsymbol{\mathcal{P}}$, the eigenvalue $k$ and the corresponding eigenfunction are updated based on a normalization condition. The fission source is then updated using the new scalar flux, and the process continues iteratively until both $k$ and $\phi(x)$ converge.

Within the MD-PNOP framework, the neural operator-based preconditioner is used to accelerate the inner loop, and the detailed formulations and derivations are presented in Appendix A2. Two different algorithms are implemented to solve the neutron transport eigenvalue problems using both DeepONet and FNO:

- Simple Preconditioning (SP): The trained neural operator directly replaces the model-based fixed source solver $\boldsymbol{\mathcal{M}}$. After convergence of the neural operator-based eigenvalue solver, the final eigenvalue $k$ and scalar flux $\phi(x)$ are used as initial guesses for the model-based eigenvalue solver for further refinement.
- Constrained Preconditioning (CP): Compared to SP, this algorithm uses the model-based solver to constrain the neural operator predictions during each inner loop iteration. After convergence of the neural operator-based solver, the final eigenvalue $k$ and scalar flux $\phi(x)$ are likewise passed as initial guesses to the model-based solver to obtain the final refined solutions.

Three test cases are presented here to demonstrate the MD-PNOP framework for eigenvalue problems. The convergence criterion of both eigenvalue $k$ and scalar flux $\phi(x)$ for all solvers is uniformly set to $10^{-4}$ for fair comparison. The performance metrics, including normalized simulation time, $L_2$-norm error, and final $k$ values, are summarized in Tables 5, 6, and 7.

Case 1 represents a fundamental baseline scenario using the same cross section parameters as the training set ($\Sigma_t^* = 1.0cm^{-1}$, $\Sigma_{s,0}^* = 0.5cm^{-1}$), but with a constant fission cross section $\Sigma_f = 0.3cm^{-1}$, as shown in Figure 6(b). This constant source distribution differs significantly from the GRF sources used during training and can be regarded as an extrapolation test case. As shown in Figure 6(a), the scalar flux solutions from all solvers match the benchmark well. However, as indicated in Table 6, the $L_2$-norm error of the FNO-SP algorithm is higher than that of the other methods. This discrepancy may arise from the nature of Fourier layers, which inherently struggle to approximate flat, constant functions accurately. As shown in Table 5, since Case 1 closely resembles the fixed source problem configurations in the training dataset, the simplest algorithm, DON-SP, achieves the best acceleration performance, requiring only 2.23% of the model-based solver's computational time while maintaining comparable accuracy for both the flux and $k$ value.

Case 2 considers a heterogeneous problem with sinusoidal spatial variations in $\Sigma_t(x)$, $\Sigma_{s0}(x)$ and $\Sigma_f(x)$, as illustrated in Figure 7(b). This heterogeneous configuration leads to irregular scalar flux distributions, as shown in Figure 7(a). Nevertheless, due to the equation reformulation technique and the hybrid design of MD-PNOP, all solvers accurately resolve the problem. Given the increased complexity of Case 2, both DeepONet and FNO achieve higher accuracy using the constrained approach, as shown in Table 6. Furthermore, in this case, the computational overhead from FFT and IFFT operations in FNO becomes negligible relative to the total computational cost, resulting in comparable total computational times for FNO-based and DeepONet-based solvers, as shown in Table 5. Overall, for the heterogeneous problem in Case 2, MD-PNOP reduces total computational time by up to 56% while fully preserving accuracy.

Case 3 is adapted from an example in the work by Nease et al. [42], representing a typical multi slab geometry, which results in step function-like cross section profiles, as shown in Figure 8(b). This system is a high-dominance ratio problem with the ratio of the first two eigenvalues of the system larger than 0.999, which significantly slows the convergence. Moreover, as shown in Figure 8(a), the second dominant eigenfunction introduces asymmetric contamination, which accumulates numerical errors during iterative use of DeepONet and ultimately leads to incorrect solutions in the DON-SP solver. The FNO-SP method performs better than DON-SP due to the global feature extraction capabilities of Fourier layers, however, it still yields a relatively large $L_2$-norm error, as indicated in Table 6. After applying constraints via the governing equations in each inner loop, both DON-CP and FNO-CP achieve the expected accuracy and reduce total computational time by up to 58%. In this complex case, FNO-CP begins to outperform DON-CP due to the superior generalization capabilities of FNO.

The above three cases collectively demonstrate the capabilities of the MD-PNOP framework under varying problem complexities. For simple and fast problems, using DON-SP can sharply reduce total computational cost without compromising accuracy. As problem complexity increases, the Constrained Preconditioning (CP) algorithm offers both high accuracy and improved computational efficiency. In complex scenarios, the overhead from FFT and IFFT operations in FNO becomes negligible, making FNO-CP the fastest and most accurate solver in Case 3. Additionally, DON-CP continues to provide fast and accurate solutions, further illustrating the architecture-agnostic nature of MD-PNOP.

Table 5. Normalized simulation time for the single group eigenvalue problems (%).

| | Model-based | DON-SP | DON-CP | FNO-SP | FNO-CP |
|---|---|---|---|---|---|
| Case 1 | 100 | 2.226 | 22.093 | 27.518 | 58.592 |
| Case 2 | 100 | 74.923 | 43.382 | 48.935 | 47.265 |
| Case 3 | 100 | 79.399 | 42.705 | 95.711 | 41.823 |

Table 6. $L_2$-norm error compared to the benchmark ($\times 10^{-4}$).

| | Model-based | DON-SP | DON-CP | FNO-SP | FNO-CP |
|---|---|---|---|---|---|
| Case 1 | / | 1.2 | 2.4 | 43.2 | 6.1 |
| Case 2 | / | 11.6 | 4.8 | 3.8 | 3.6 |
| Case 3 | / | 274.9 | 6.3 | 147.8 | 6.9 |

Table 7. Eigenvalue $k$ from different solvers.

| | Model-based | DON-SP | DON-CP | FNO-SP | FNO-CP |
|---|---|---|---|---|---|
| Case 1 | 1.71402 | 1.71444 | 1.71449 | 1.71507 | 1.71455 |
| Case 2 | 0.40253 | 0.40257 | 0.40255 | 0.40255 | 0.40255 |
| Case 3 | 0.33258 | 0.33261 | 0.33258 | 0.33258 | 0.33258 |

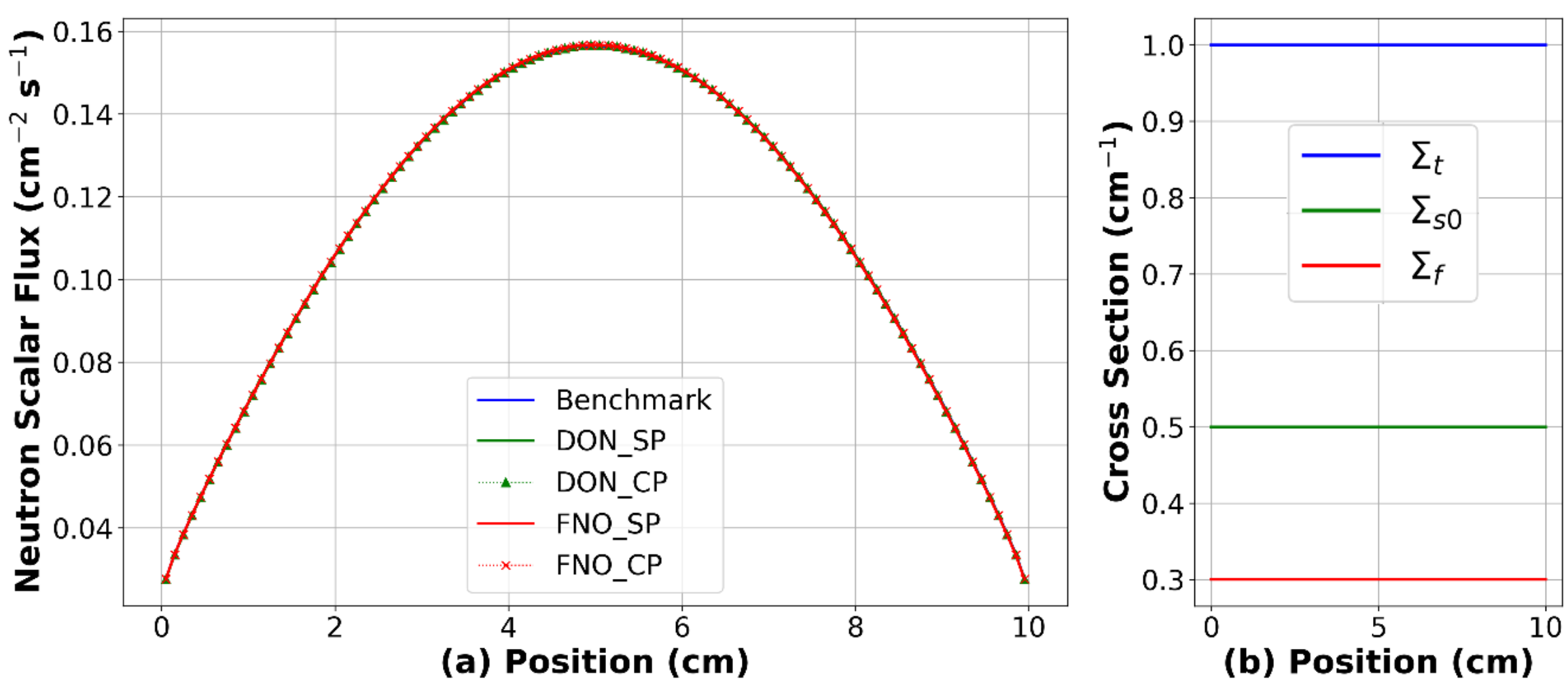

Figure 6: Testing case 1 for the single group eigenvalue problem. (a) Neutron scalar flux comparison of different solvers. (b) The cross sections used for Case 1. All three cross sections are constants.

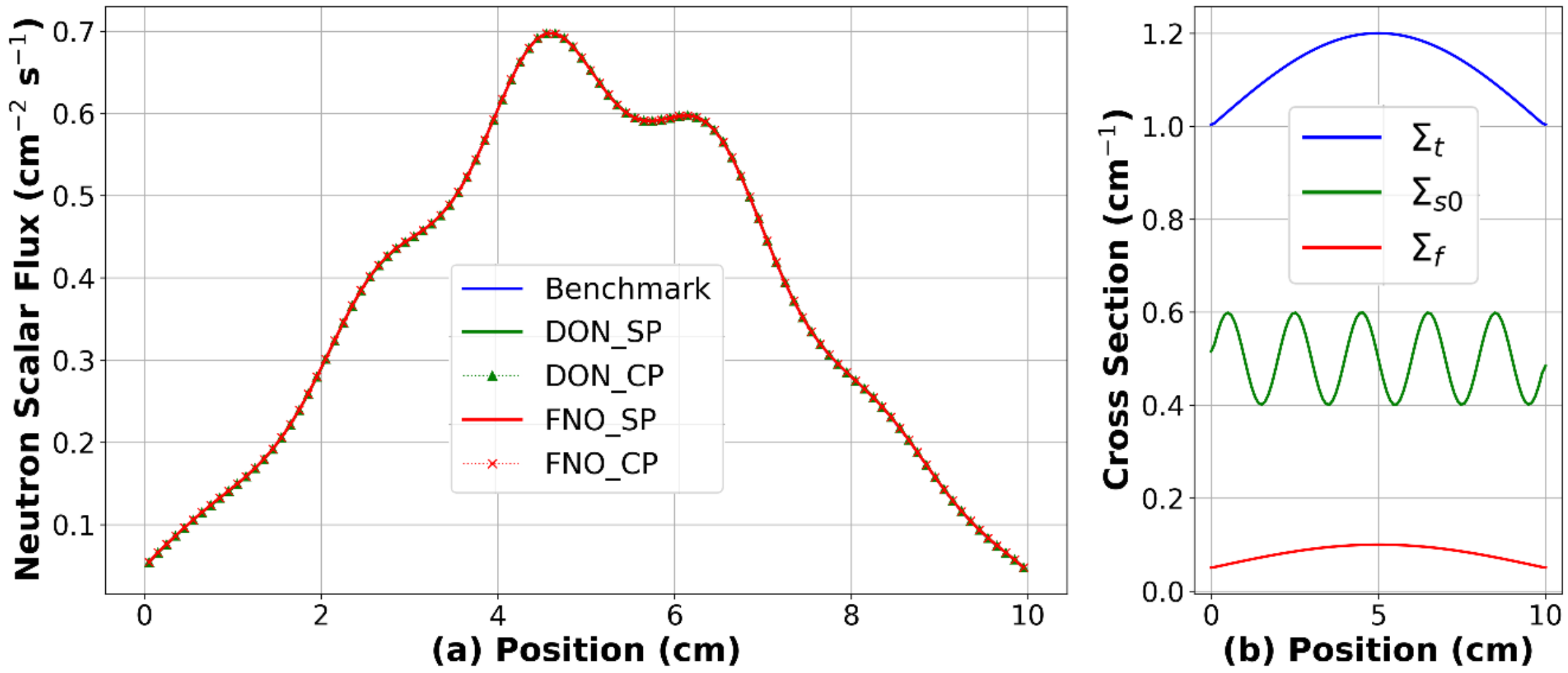

Figure 7: Testing case 2 for the single group eigenvalue problem. (a) Neutron scalar flux comparison of different solvers. (b) The cross sections used for Case 2. All three cross sections are sinusoidal functions.

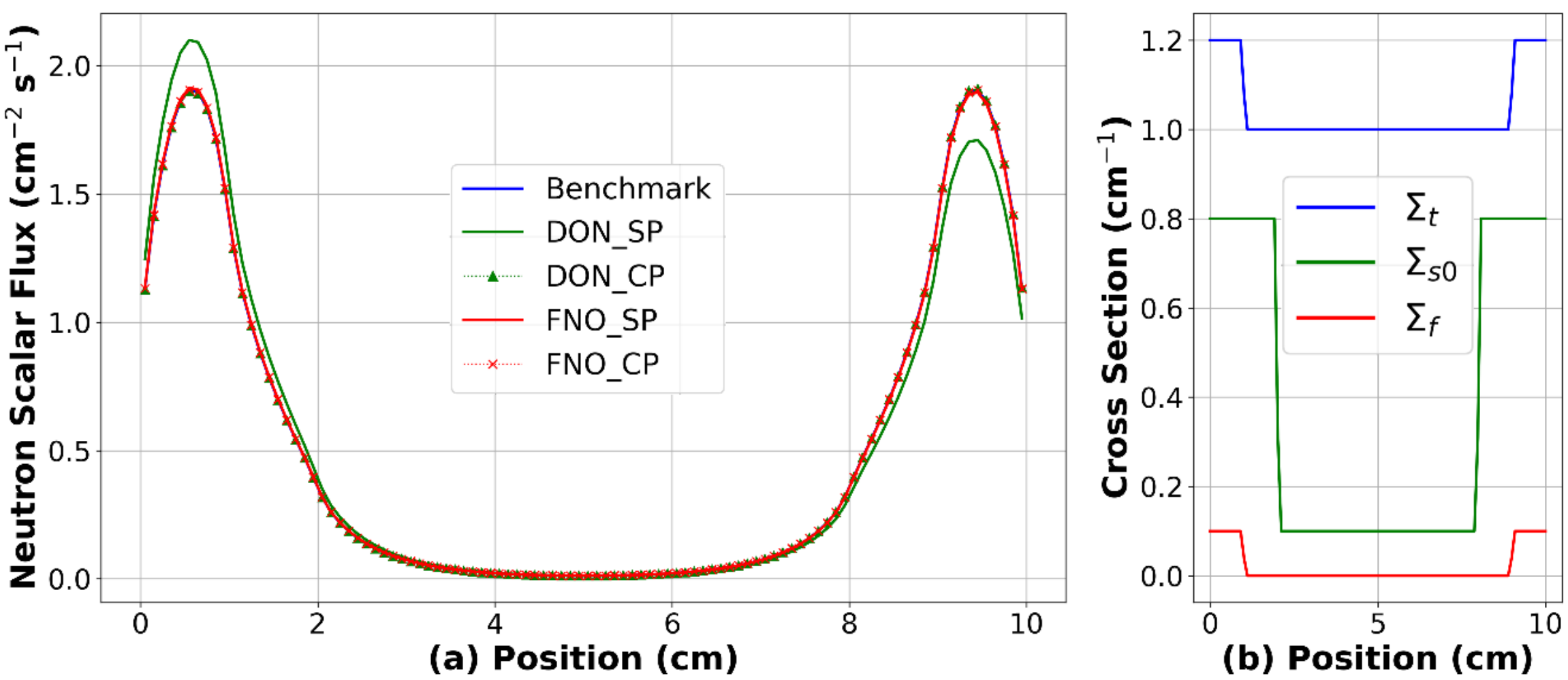

Figure 8: Testing case 3 for the single group eigenvalue problem, which is a multi-slab geometry adopted from an example in the work by Nease et al. [42]. (a) Neutron scalar flux comparison of different solvers. (b) The cross sections used for Case 3. All three cross sections are step functions.

### 5.3 Multigroup Eigenvalue Problem

In nuclear reactor design and criticality calculations, neutrons are typically grouped into different energy ranges to account for their energy-dependent cross sections, which lead to distinct neutron behaviors [43]. Consequently, multigroup neutron transport eigenvalue problems are widely used in practical applications. A representative three-group system can be formulated as follows, where upscattering (from lower to higher energies) and anisotropic scattering are neglected:

Group 1 (fast neutron, highest energy):

$$\mu\frac{\partial\psi_1(x,\mu)}{\partial x}+\Sigma_{t,1}(x)\psi_1(x,\mu)=\frac{1}{2}\Sigma_{s,0}^{1\to1}(x)\phi_1(x)+\frac{\chi_1}{2k}\sum_{g'=1}^{3}\nu_{g'}\Sigma_{f,g'}\phi_{g'}(x) \tag{27}$$

Group 2 (intermediate energy neutron):

$$\begin{aligned}\mu\frac{\partial\psi_2(x,\mu)}{\partial x}&+\Sigma_{t,2}(x)\psi_2(x,\mu)\\&=\frac{1}{2}\Sigma_{s,0}^{1\to2}(x)\phi_1(x)+\frac{1}{2}\Sigma_{s,0}^{2\to2}(x)\phi_2(x)+\frac{\chi_2}{2k}\sum_{g'=1}^{3}\nu_{g'}\Sigma_{f,g'}\phi_{g'}(x)\end{aligned} \tag{28}$$

Group 3 (thermal neutron, lowest energy):

$$\mu\frac{\partial\psi_3(x,\mu)}{\partial x}+\Sigma_{t,3}(x)\psi_3(x,\mu)=\frac{1}{2}\sum_{g'=1}^{3}\Sigma_{s,0}^{g'\to3}(x)\phi_{g'}(x)+\frac{\chi_3}{2k}\sum_{g'=1}^{3}\nu_{g'}\Sigma_{f,g'}\phi_{g'}(x) \tag{29}$$

where the subscript $g$ denotes the energy group, and the fission spectrum $\chi_g$ represents the fraction of fission neutrons emitted into the energy group $g$. Consequently, these three equations exchange neutrons as additional source terms via scattering and fission reactions, forming a tightly coupled system of transport equations. In such complex systems, conventional neural network-based solvers would require extensive retraining to accommodate different parameter configurations, making them impractical. In contrast, the MD-PNOP framework can handle these scenarios effectively using the same neural operators trained on minimal datasets.

For demonstration purposes and to emulate the homogenized reactor pin cell configurations, all cross sections for the multigroup problem are set as constants. It is important to note that, as demonstrated in the single-group eigenvalue cases, there is no fundamental limitation in the MD-PNOP framework that would prevent its extension to spatially heterogeneous distributions. The relevant parameters used are summarized in Tables 8 and 9. Since all cross sections are constant, only the SP algorithm is demonstrated here, implemented using both DeepONet and FNO architectures.

Table 8. Scattering cross sections for the three-group eigenvalue problem.

| $\Sigma_{s,0}^{g\to g'}(cm^{-1})$ | $g' = 1$ | $g' = 2$ | $g' = 3$ |
|---|---|---|---|
| $g = 1$ | 0.024 | 0.171 | 0.033 |
| $g = 2$ | 0.000 | 0.600 | 0.275 |
| $g = 3$ | 0.000 | 0.000 | 2.000 |

Table 9. Parameters for the three-group eigenvalue problem.

| | $\nu$ | $\Sigma_{t,g}(cm^{-1})$ | $\Sigma_{f,g}(cm^{-1})$ | $\chi_g$ |
|---|---|---|---|---|
| $g = 1$ | 3.0 | 0.240 | 0.006 | 0.96 |
| $g = 2$ | 2.5 | 0.975 | 0.060 | 0.04 |
| $g = 3$ | 2.0 | 3.000 | 0.900 | 0.00 |

The results of different solvers are summarized in Table 10 and illustrated in Figure 9. Both DON-SP and FNO-SP accurately solve the multigroup eigenvalue problem and reduce the total computational time. Notably, DON-SP achieves more than a 50% reduction in computational cost compared to the model-based solver. These findings demonstrate that, for complex and tightly coupled multigroup systems, the MD-PNOP framework exhibits robust performance across both neural operator architectures. This capability is particularly beneficial for multiphysics systems [44-46], which typically involve intricate couplings among multiple governing equations and are therefore computationally demanding to solve. Within the MD-PNOP framework, multiple neural operators can be trained independently for individual equations, with system-level predictions obtained through the coupled neural operator ensemble. These predictions serve as enhanced initial guesses for the model-based solver, enabling refinement to fully satisfy physical constraints. Consequently, the present results affirm the effectiveness of the proposed MD-PNOP framework and highlight its potential to accelerate PDE solvers, offering significant benefits for a wide range of practical engineering applications.

Table 10. Results of the three-group eigenvalue problem using different solvers.

|  | Model-based | DON-SP | FNO-SP |
|---|---|---|---|
| Normalized Time (%) | 100 | 46.109 | 76.619 |
| Eigenvalue $k$ | 1.30621 | 1.30611 | 1.30597 |

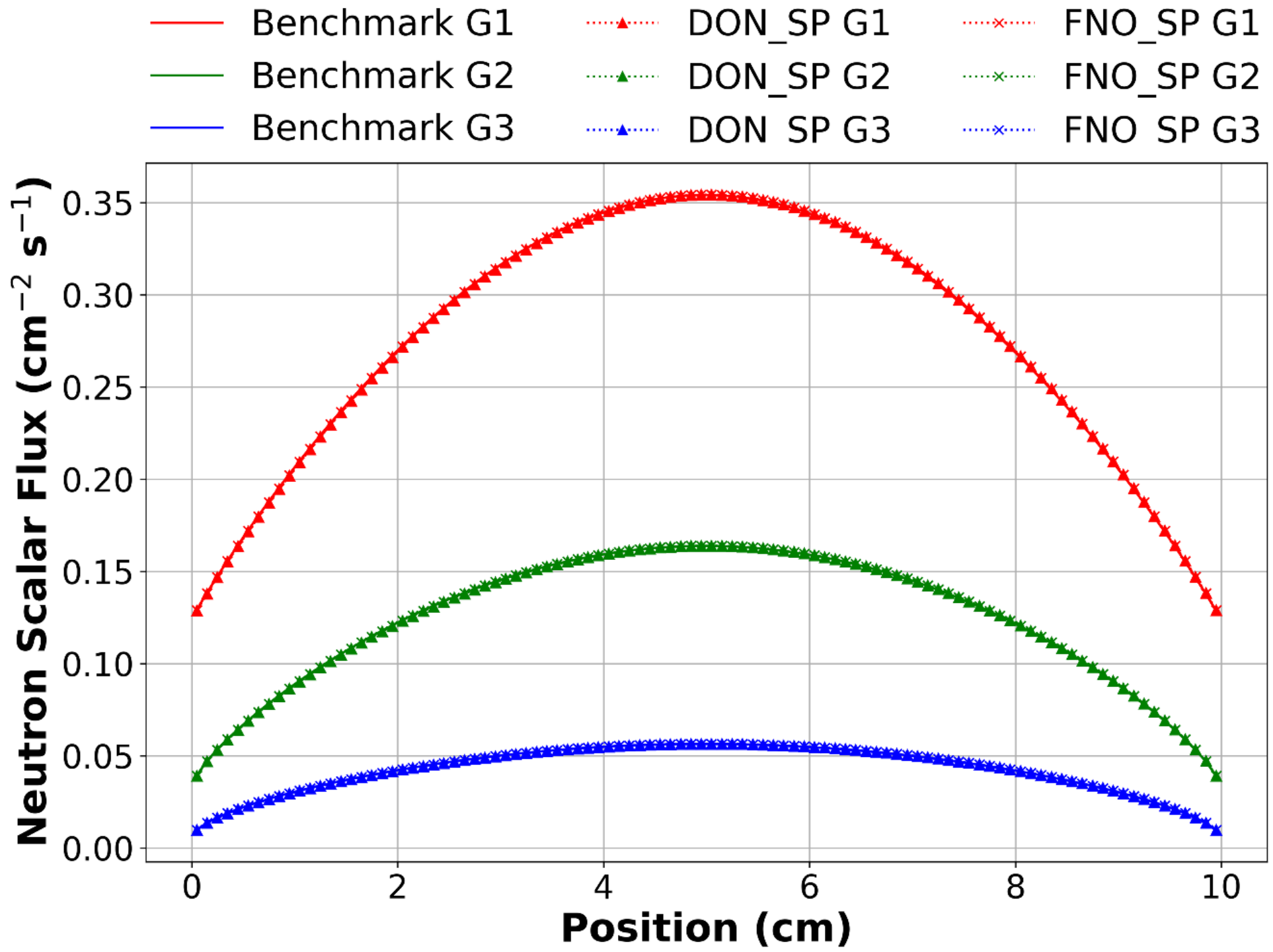

Figure 9: Neutron scalar flux distributions for the three-group eigenvalue problem using different solvers. The results are shown for each energy group: Group 1 (red), Group 2 (green), and Group 3 (blue). Solid lines represent benchmark solutions; dotted lines with symbols correspond to predictions from DON-SP and FNO-SP algorithms.

## 6. Conclusion

In this work, we developed the Minimal-Data Parametric Neural Operator Preconditioning (MD-PNOP) framework to accelerate solvers for partial differential equations (PDEs). MD-PNOP offers a generalizable approach to integrate neural operators into traditional numerical solvers while maintaining full physics constraints and full-order accuracy. By design, MD-PNOP inherently addresses black-box concerns, mitigates the typical accuracy degradation and interpretability issues associated with purely data-driven neural solvers.

The MD-PNOP framework integrates two central innovations: an equation recast strategy and a hybrid solver architecture. The recast approach, inspired by perturbation theory, reformulates parameter deviations from the training configuration as additional source terms. This recast enables neural operators trained on minimal datasets to generalize effectively to previously unseen parameter regimes without retraining. The hybrid solver then leverages the neural operator's predictions as high-quality initial guesses, reducing the convergence time of iterative solvers. By subsequently refining the neural operator's predictions through a full-order model-based solver, MD-PNOP guarantees that final solutions are fully accurate and rigorously satisfy the governing PDEs. Furthermore, two algorithmic variants, Simple Preconditioning (SP) and Constrained Preconditioning (CP), provide flexibility to handle eigenvalue problems of varying complexity.

MD-PNOP was thoroughly evaluated and demonstrated by solving the neutron transport equation under diverse scenarios, including fixed source problems, single-group eigenvalue problems, and multigroup coupled eigenvalue problems. The neural operators were trained using only a single set of constant cross sections, yet MD-PNOP successfully handled constant, heterogeneous sinusoidal, and even step-function cross section distributions. Across all test cases, MD-PNOP consistently maintained high accuracy while reducing total computational cost, achieving over 50% reduction in simulation time in most scenarios.

The implementation of MD-PNOP using both Deep Operator Network (DeepONet) and Fourier Neural Operator (FNO) further demonstrates its architecture-agnostic nature, providing flexibility to adopt or incorporate future neural operator advancements. Observations from this study suggest that DeepONet offers extremely fast inference and is particularly suitable for simpler, homogeneous problems where maximal acceleration is desired. In contrast, FNO benefits from its efficient training and generally stronger generalization performance, making it well-suited for more complex, heterogeneous problems. These insights provide practical guidance for selecting appropriate neural operator architectures when applying MD-PNOP to domain-specific problems.

Since the equation-recast formulation is defined at the operator level and does not rely on dimension-specific assumptions, the overall structure of the hybrid solver and the associated preconditioning strategy remains unchanged when extended to higher-dimensional problems. The neural operator backbones employed are also naturally extensible to higher spatial dimensions [47]. The primary challenge in higher dimensions arises from the numerical approximation involved in the practical evaluation of the operator difference. As discussed in Section 3.3, such approximation errors primarily affect the efficiency of the preconditioning rather than the correctness of the final solution. Importantly, the equation-recast framework allows selective or partial recasting of parameter dependencies, providing a flexible mechanism to control operator complexity and mitigate numerical approximation errors in high-dimensional settings. These features suggest that MD-PNOP can be effectively extended to higher-dimensional problems, particularly in scenarios where accelerating high-fidelity solvers while preserving physical consistency is of primary importance.

Overall, MD-PNOP represents a promising strategy for combining data-driven approaches with model-based numerical solvers, offering a robust, efficient, and physically consistent framework for accelerating PDE solvers across a wide range of engineering and scientific applications.

## Appendix

### A1. Training Data and Neural Operator Details

The training dataset used in this work consists of 1,000 samples of neutron source and scalar flux pairs. For both the neutron source and scalar flux, a spatial discretization of $\Delta L = 0.1cm$ is employed in a slab geometry of length $L = 10cm$, resulting in 100 center points. The neutron sources are sampled from a Gaussian Random Field (GRF) with a mean of 0.07, variance of 0.0003, and length scale of 1.2. After sampling, the sources are normalized to represent a single neutron (the integral of each source function over the spatial domain equals unity), ensuring scalability and consistency across different test cases.

Since the MD-PNOP framework is designed to be agnostic to the choice of neural operator architecture, two widely used and representative neural operator models, namely the Deep Operator Network (DeepONet) and the Fourier Neural Operator (FNO), are selected as backbones for demonstration purposes. The use of these two architectures is intended to illustrate the generality and robustness of the proposed framework across distinct neural operator designs, rather than to benchmark or optimize the performance of specific architectures. Accordingly, only limited and coarse grid scans over key architectural hyperparameters are conducted to ensure stable training behavior and to achieve comparable levels of training loss that are sufficient for providing reliable baseline predictions within the MD-PNOP framework. No extensive hyperparameter optimization or state-of-the-art architecture tuning is performed. As a result, the reported training costs and standalone prediction accuracies of DeepONet and FNO should not be interpreted as a direct performance comparison between the two neural operators.

Both neural operator models are implemented using the PyTorch library [48], optimized with the Adam optimizer [49] and trained using a mean squared error loss. Training is conducted on a desktop workstation equipped with an NVIDIA RTX 4090 GPU and an Intel i9-14900KS CPU. The final architectural configurations and training hyperparameters adopted in this work are summarized in Table A1.

The DeepONet used in this study consists of a Branch Network (BN) and a Trunk Network (TN). The BN is a fully connected neural network (FNN) with architecture $[100,200,200,100]$. BN takes the discretized neutron source $S(x)$ at 100 spatial points as input and outputs a 100-dimensional coefficient vector. The TN is also an FNN with architecture $[1,200,200,100]$, taking the spatial coordinate $x$ as input and producing a 100-dimensional basis vector. The final DeepONet output is obtained via the inner product of the BN and TN outputs, yielding the scalar flux $\phi(x)$ for a given source $S$. ReLU activation functions are used throughout the network. The selected BN and TN architectures are based on a limited grid scan over the number of layers and hidden-layer widths, together with insights from prior characterization studies [18]. The resulting DeepONet contains a total of $141{,}200$ trainable parameters. Training was carried out for up to 10,000 epochs, corresponding to a wall-clock time exceeding 20 hours on the specified hardware. DeepONet achieved the best training loss of $8 \times 10^{-8}$.

The Fourier Neural Operator (FNO) architecture takes the discretized neutron source $S(x)$ and the corresponding spatial coordinates as input. Each input pair $(S(x_i), x_i)$ is first lifted to a 64-dimensional feature space using a fully connected layer. The lifted features are then processed through four standard Fourier layers, each consisting of a spectral convolution implemented via FFT-IFFT operations and a parallel pointwise linear transformation in physical space. The number of retained Fourier modes is truncated to 16 to filter high-frequency components, and the final output is projected back onto the spatial grid to obtain the predicted scalar flux field. ReLU activation functions are used throughout the FNO. The adopted configuration is selected based on a limited grid scan over the number of Fourier layers, the number of retained modes, and the lifting dimension, balancing representational capacity and computational efficiency. The resulting FNO contains 549,569 trainable parameters. Training completes in approximately 7 minutes over 10,000 epochs, achieving a best training loss of $2 \times 10^{-8}$.

The substantial difference in training time between DeepONet and FNO primarily stems from their distinct architectural characteristics and data processing strategies. DeepONet is inherently mesh-free and pointwise: for each training sample, predictions are evaluated independently at all spatial locations through the trunk network and combined with the branch network outputs. Consequently, each training sample effectively corresponds to multiple forward evaluations (100 in the present setting), leading to a higher computational cost per training epoch. In contrast, FNO operates directly on discretized fields and processes the entire spatial domain simultaneously using FFT-based global convolutions, which are highly optimized for GPU execution and enable efficient batch processing despite the larger number of trainable parameters. Accordingly, the two neural operators are designed to target different use cases: DeepONet emphasizes mesh-free representations suitable for continuous or irregular evaluation points, whereas FNO operates on fixed discretized grids with efficient spectral convolutions. As a result, their respective advantages lie in different aspects of the operator learning problem. Therefore, the reported training costs and standalone prediction accuracies should not be interpreted as a direct performance comparison between DeepONet and FNO, but rather as evidence that MD-PNOP can consistently leverage different neural operator backbones under comparable and transparent training settings.

Table A1. Neural operator architectures and training hyperparameters.

| **Category** | **DeepONet** | **FNO** |
|---|---|---|
| Input | $S(x), x_i$ | $S(x), x$ |
| Output | $\phi(x_i)$ | $\phi(x)$ |
| Architecture | BN: FNN [100,200,200,100]<br>TN: FNN [1,200,200,100] | No. of Fourier layers: 4<br>Fourier mode: 16<br>Lifting width: 64 |
| Activation | ReLU | ReLU |
| Optimizer | Adam | Adam |
| Loss function | Mean Squared Error (MSE) | Mean Squared Error (MSE) |
| Initial learning rate | $1 \times 10^{-3}$ | $1 \times 10^{-3}$ |
| Batch size | 100 | 32 |
| No. Trainable parameters | 141,200 | 549,569 |
| Training time (10,000 epochs) | ~20 hours | ~7 minutes |
| Best training loss | $8 \times 10^{-8}$ | $2 \times 10^{-8}$ |

### A2. Eigenvalue Problem

Eigenvalue problems are fundamental in many scientific and engineering applications. Notable examples include the determination of the effective multiplication factor $k$ in nuclear reactor design and the computation of natural frequencies and mode shapes in structural mechanics for vibration analysis.

A general eigenvalue problem with the eigenvalue $\lambda$ and a parameterized operator $\mathcal{L}(\boldsymbol{p})$ can be expressed as:

$$\mathcal{L}(\boldsymbol{p})[u(\boldsymbol{x})] = \lambda B(\boldsymbol{q})[u(\boldsymbol{x})] \tag{A1}$$

where $B(\boldsymbol{q})$ is an operator applied to the eigenfunction, parameterized by $\boldsymbol{q}$, resulting in a source-like term.

In practice, classical numerical eigenvalue solvers consist of two nested iterative procedures:

- an inner iteration operator $\boldsymbol{\mathcal{M}}$, which solves the prescribed source problem using the current eigenvalue and eigenfunction estimates;
- an outer update operator $\boldsymbol{\mathcal{P}}$, which calculates the eigenvalue, normalizes the eigenfunction, and then updates the source term.

Given the current eigenvalue estimate $\lambda^{(n)}$, the fixed-source problem is solved via the nonlinear iteration operator $\boldsymbol{\mathcal{M}}$:

$$u^{(n+1/2)}(\boldsymbol{x}) = \boldsymbol{\mathcal{M}}\left(u^{(n)}(\boldsymbol{x}), \lambda^{(n)} B(\boldsymbol{q})\left[u^{(n)}(\boldsymbol{x})\right]; \boldsymbol{p}\right) \quad \text{(A2)}$$

The inner solver corresponds to repeated applications of $\boldsymbol{\mathcal{M}}$ until convergence for the current eigenvalue estimate. After convergence of the inner iteration, the eigenvalue is updated using a Rayleigh quotient or a normalization condition, and the eigenfunction is subsequently normalized according to a problem-specific rule (e.g., $\|u\| = 1$). Thus, in the outer iteration loop $\boldsymbol{\mathcal{P}}$, both eigenvalue and eigenfunction are updated before proceeding to the next inner iteration:

$$u^{(n+1)}(\boldsymbol{x}), \lambda^{(n+1)} = \boldsymbol{\mathcal{P}}\left(u^{(n+1/2)}\right) \quad \text{(A3)}$$

The updated pair $\left(\lambda^{(n+1)}, u^{(n+1)}\right)$

$$u^{(n+1+1/2)}(\boldsymbol{x}) = \boldsymbol{\mathcal{M}}\left(u^{(n+1)}(\boldsymbol{x}), \lambda^{(n+1)} B(\boldsymbol{q})\left[u^{(n+1)}(\boldsymbol{x})\right]; \boldsymbol{p}\right) \quad \text{(A4)}$$

The convergence criteria of eigenvalue problems include the convergence of the field function $u^{(n)}(\boldsymbol{x})$ in the inner loop, and the simultaneous convergence of both the eigenvalue $\lambda$ and the final normalized eigenfunction $u(x)$ in the outer iterations. While updating the eigenvalue itself (outer iteration) is computationally inexpensive, the dominant computational cost arises from repeatedly solving the fixed source problem. Consequently, the overall computational effort scales with the number of inner iterations multiplied by the number of outer iterations. Therefore, leveraging a neural operator to accelerate the fixed source problem solver can substantially reduce the overall computational time required for eigenvalue calculations.

We developed two algorithms to accelerate the eigenvalue calculation using a neural operator trained on minimal data.

(i) Simple Preconditioning

In this approach, the neural operator replaces the fixed-source iteration operator $\boldsymbol{\mathcal{M}}$ in the inner loop:

$$u^{(n+1/2)}(\boldsymbol{x}) = \mathcal{L}_{NO}^{-1}(\boldsymbol{p}^*)\left[\lambda^{(n)} B(\boldsymbol{q})\left[u^{(n)}(\boldsymbol{x})\right] - \delta\mathcal{L}(\Delta\boldsymbol{p})\left[u^{(n)}(\boldsymbol{x})\right]\right] \quad \text{(A5)}$$

After convergence of the outer loop, the final eigenvalue and eigenfunction obtained from the neural operator are used as initial guesses for a model-based eigenvalue solver to further refine the solution:

$$u^{(1/2)}(\boldsymbol{x}) = \boldsymbol{\mathcal{M}}(u_{NO}(\boldsymbol{x}), \lambda_{NO} B(\boldsymbol{q})[u_{NO}(\boldsymbol{x})]; \boldsymbol{p}) \quad \text{(A6)}$$

(ii) Constrained Preconditioning

In this variant, the neural operator acts as a preconditioner within each inner iteration step. First, a neural-operator prediction is obtained:

$$u^{(n+1/4)}(\boldsymbol{x}) = \mathcal{L}_{NO}^{-1}(\boldsymbol{x}; \boldsymbol{p}^*)\left[\lambda^{(n)} B(\boldsymbol{p})\left[u^{(n)}(\boldsymbol{x})\right] - \delta\mathcal{L}(\Delta\boldsymbol{p})\left[u^{(n)}(\boldsymbol{x})\right]\right] \quad \text{(A7)}$$

This intermediate approximation is then refined using the classical iteration operator:

$$u^{(n+1/2)}(\boldsymbol{x}) = \boldsymbol{\mathcal{M}}\left(u^{(n+1/4)}(\boldsymbol{x}), \lambda^{(n)} B(\boldsymbol{p})\left[u^{(n+1/4)}(\boldsymbol{x})\right]; \boldsymbol{p}\right) \tag{A8}$$

This hybrid strategy enforces the governing equations at each inner iteration step, improving robustness for heterogeneous or non-smooth parameter distributions. Although additional model-based updates reduce the theoretical speed-up compared to the simple preconditioning strategy, stability is enhanced. After convergence of the outer loop, the final eigenvalue and eigenfunction obtained from the neural operator are still used as initial guesses for a model-based eigenvalue solver to further refine the solution.

In both strategies, the final neural-operator prediction is used as an initial approximation for a classical model-based eigenvalue solver. This final correction guarantees consistency with the governing equations and eliminates residual approximation errors introduced by operator reformulation or neural approximation. The MD-PNOP framework therefore accelerates the classical eigenvalue solver by enhancing the inner iteration operator, while preserving physical fidelity and numerical reliability.